\documentclass[journal]{vgtc}                     

\onlineid{TBD}

\vgtccategory{System \& Rendering}

\title{SciVisAgentBench: A Benchmark for Evaluating Scientific Data Analysis and Visualization Agents}

\author{%
  \authororcid{Kuangshi Ai}{0009-0005-7171-6529},
  \authororcid{Haichao Miao}{0000-0001-6580-2918},
  \authororcid{Kaiyuan Tang}{0009-0001-3512-0112},
  \authororcid{Nathaniel Gorski}{0009-0001-8205-5640},
  \authororcid{Jianxin Sun}{0000-0002-9627-9397},
  \authororcid{Guoxi Liu}{0000-0002-8164-7185},\\
  \authororcid{Helgi I.\ Ing{\'o}lfsson}{0000-0002-7613-9143},
  \authororcid{David Lenz}{0000-0002-2587-2783},
  \authororcid{Hanqi Guo}{0000-0001-7776-1834},
  \authororcid{Hongfeng Yu}{0000-0002-0596-8227},
  Teja Leburu,
  \authororcid{Michael Molash}{0009-0008-5025-6529},\\
  \authororcid{Bei Wang}{0000-0002-9240-0700},
  \authororcid{Tom Peterka}{0000-0002-0525-3205},
  \authororcid{Chaoli Wang}{0000-0002-0859-3619}, and
  \authororcid{Shusen Liu}{0000-0002-6455-8391}
}

\authorfooter{
  \item K.\ Ai, K.\ Tang, and C.\ Wang are with the University of Notre Dame.
  E-mail: \{kai, ktang2, chaoli.wang\}@nd.edu.
  \item H.\ Miao, H.\ I.\ Ing{\'o}lfsson, and S.\ Liu are with Lawrence Livermore National Laboratory.
  E-mail: \{miao1, ingolfsson1, liu42\}@llnl.gov.
  \item N.\ Gorski and B.\ Wang are with the University of Utah.
  E-mail: \{gorski, beiwang\}@sci.utah.edu.
  \item J.\ Sun and H.\ Yu are with the University of Nebraska–Lincoln.
  E-mail: \{s-jsun5, hfyu\}@unl.edu.
  \item G.\ Liu and H.\ Guo are with The Ohio State University.
  E-mail: \{liu.12722, guo.2154\}@osu.edu.
  \item D.\ Lenz and T.\ Peterka are with Argonne National Laboratory.
  E-mail: dlenz@anl.gov, tpeterka@mcs.anl.gov.
  \item T.\ Leburu and M. Molash are with Anthropic PBC. 
  E-mail: \{teja, mmolash\}@anthropic.com.  
}

\abstract{
Recent advances in large language models (LLMs) have enabled agentic systems to translate natural-language intent into executable scientific visualization (SciVis) tasks. Despite rapid progress, the community lacks a principled and reproducible benchmark for evaluating these emerging SciVis agents in realistic, multi-step analysis settings. We present SciVisAgentBench, a comprehensive and extensible benchmark for evaluating scientific data analysis and visualization agents. Our benchmark is grounded in a structured taxonomy spanning four dimensions: application domain, data type, complexity level, and visualization operation. It currently comprises 108 expert-crafted cases covering diverse SciVis scenarios. To enable reliable assessment, we introduce a multimodal outcome-centric evaluation pipeline that combines LLM-based judging with deterministic evaluators, including image-based metrics, code checkers, rule-based verifiers, and case-specific evaluators. We also conduct a validity study with 12 SciVis experts to examine the agreement between human and LLM judges. Using this framework, we evaluate representative SciVis agents and general-purpose coding agents to establish initial baselines and reveal capability gaps. SciVisAgentBench is designed as a living benchmark to support systematic comparison, diagnose failure modes, and drive progress in agentic SciVis. The benchmark is available at \url{https://scivisagentbench.github.io/}.
}


\graphicspath{{figs/}{figures/}{pictures/}{images/}{./}} 

\usepackage{tabu}                      
\usepackage{booktabs}                  
\usepackage{lipsum}                    
\usepackage{mwe}                       

\usepackage{mathptmx}                  
\usepackage{amsmath}
\usepackage{comment}
\usepackage{longtable,booktabs,array}
\usepackage{multirow}
\usepackage{adjustbox}

\newenvironment{myitemize}{
\begin{itemize}
 \setlength{\itemsep}{1pt}
 \setlength{\parskip}{0pt}
 \setlength{\parsep}{0pt}}{\end{itemize}
}

\newcommand{\hot}[1]{{#1}}

\begin{document}


\firstsection{Introduction}
\maketitle
Recent advances in large language models (LLMs) and multimodal LLMs (MLLMs) have given rise to \emph{agentic visualization} systems, in which (semi-)autonomous agents translate natural language intent into executable visualization actions~\cite{dhanoa2025agentic, liu2024ava, liu2025paraview, ai2025nli4volvis}. In scientific visualization (SciVis), such agents promise to assist domain experts in complex exploratory workflows and lower the barrier to accessing sophisticated visualization tools for non-expert users. A growing body of agents, including AVA~\cite{liu2024ava}, ChatVis~\cite{peterka2025chatvis}, and ParaView-MCP~\cite{liu2025paraview}, together with agentic systems such as NLI4VolVis~\cite{ai2025nli4volvis}, VizGenie~\cite{biswas2025vizgenie}, and InferA~\cite{tam2025infera}, show the feasibility of tool-using, multimodal, and 
\hot{multi-step} visualization agents operating over scientific data. As these systems rapidly mature, a central question emerges: \emph{how should progress in SciVis agents be measured in a principled, reproducible, and scalable way}?

Historically, many significant advances in machine learning have been catalyzed by the availability of strong, widely adopted benchmarks. ImageNet~\cite{deng2009imagenet}, for example, provides a common ``yardstick'' that enables systematic comparison, reveals failure modes, and ultimately accelerates the deep learning revolution. In contrast, despite the growth of SciVis agents, the community lacks a comparable measurement framework. While numerous benchmarks exist for general-purpose agents and for {\em natural-language-to-visualization} (NL2VIS) or chart generation tasks (e.g., NL4DV~\cite{Narechania-TVCG21} and VisEval~\cite{chen2024viseval}), their scope remains limited. These benchmarks typically focus on short-horizon tasks, 2D charts, or declarative plotting, and do not capture the defining characteristics of SciVis: high-dimensional data, view-dependent semantics, complex pipelines composed of multiple operations, and \hot{structured multi-step visualization workflows}. Consequently, existing evaluations fall short of providing a reliable benchmark for measuring agent capability in scientific data analysis and visualization. 

Evaluating visualization agents is intrinsically challenging for SciVis. 
First, meaningful SciVis tasks frequently require multi-step workflows involving data loading, filtering, parameter tuning, rendering choices, and view manipulation, rather than single-shot plot generation. 
Second, visualization tasks often admit multiple valid outcomes, making the specification of ground truth inherently difficult. 
Third, the notion of ``correctness'' is often entangled with subjective judgment and domain expertise, complicating objective evaluation. 
These issues are already non-trivial for visualization benchmarks in general, but become even more pronounced in SciVis, where {\em scientific insights}, rather than {\em visual appearances} alone, are the ultimate goal. A natural concern, therefore, is that any benchmark risks being overly general given the diversity of SciVis applications and domains. This concern should not preclude benchmarking itself; instead, it motivates a design that emphasizes extensibility, expert grounding, and long-term growth.

To address these challenges, we present \textbf{SciVisAgentBench}, a comprehensive and extensible {\bf Bench}mark for evaluating {\bf Sci}entific data analysis and {\bf Vis}ualization {\bf Agent}s. The benchmark is grounded in a taxonomy of SciVis domains, datasets, and tasks. It is constructed in close collaboration with SciVis experts across canonical visualization areas, including volume, flow, biomedical image, molecular visualization, and topological data analysis. These areas are by no means inclusive, but provide a good starting point for us to gather cases. Experts curate and design representative tasks within their subfields, resulting in a collection of 108 cases of varying difficulty. While the benchmark does not exhaustively cover the entire taxonomy, it is designed as a starting point for a living framework: new agents can be evaluated with minimal integration effort, and researchers can contribute additional domain-specific tasks, datasets, and evaluators as the field evolves.

Because image outputs from SciVis tasks differ from those of multiple-choice items in other benchmarks and evaluations, we validate SciVisAgentBench through a multifaceted validation study. We begin by assessing the capability of our LLM judge by having it evaluate agent-generated visualizations against expert-defined ground truths using structured, expert-authored rubrics. We conduct a human-LLM alignment study, in which SciVis experts and an MLLM judge independently evaluate visualization outcomes, enabling us to quantify agreement and reliability. To further stress-test the evaluation process, we examine the robustness of the LLM judge across variations in prompts and presentation, as well as its limitations in multimodal understanding. Using this framework, we evaluate a range of existing SciVis agents and capable general-purpose agents across multiple backbone LLMs and report their performance to establish initial baselines for the community.

\hot{While real-world SciVis tasks often involve open-ended goals, long-horizon exploration, and multiple valid visualization outcomes, these settings remain difficult to evaluate reproducibly at a benchmark scale. Therefore, SciVisAgentBench focuses on structured tasks with explicit and verifiable outcomes. We view this benchmark as a foundational first step toward broader evaluation of open-ended, long-horizon scientific data analysis agents.} 
This paper makes the following contributions:
\begin{myitemize} 
\vspace{-0.05in}
\item{We introduce an easy-to-use, extensible benchmark framework that enables systematic evaluation of SciVis agents operating over real-world scientific tools and data.}
\item{We propose a multimodal outcome-based evaluation pipeline that combines LLM-based judging with rule-based and deterministic evaluators, and analyze its reliability for SciVis assessment.}
\item{To provide a comprehensive coverage of SciVis tasks, we present a structured taxonomy of the SciVis domain, data, and tasks that guides benchmark design and coverage.}
\item {We release the benchmark, including datasets, tasks, ground-truth artifacts, and evaluation rubrics spanning multiple domains and difficulty levels, and invite community contributions to support its continued evolution.}
\vspace{-0.05in}
\end{myitemize}





	
\vspace{-0.075in}
\section{Related Work}

Evaluating AI agents has become increasingly important as systems evolve from static language models to multimodal, tool-using agents with long-horizon interaction. Most existing benchmarks focus on web automation, code execution, or conversational tasks, with limited attention to scientific data analysis and visualization. SciVis introduces challenges such as iterative exploration and domain-aware interpretation that are poorly captured by these settings. Accordingly, we review prior work spanning general agent evaluation, agentic visualization systems, and visualization-specific benchmarks.

\vspace{-0.05in}
\subsection{General Agent Evaluation Frameworks}

\textbf{Comprehensive agent benchmarks.}
The broader AI community has proposed several general-purpose frameworks for evaluating LLM-based agents. AgentBench~\cite{liu2023agentbench} and AgentBoard~\cite{ma2024agentboard} offer standardized testbeds for multi-turn agents, focusing on task completion and reasoning accuracy. While these benchmarks establish proper infrastructure, they treat visualization as a generic task and do not account for its exploratory, iterative, and perceptual characteristics. Kapoor et al.\ \cite{kapoor2024ai} critiqued current evaluation practices, advocating for joint optimization of accuracy and cost metrics rather than focusing solely on task completion rates.
Other benchmarks emphasize interaction and robustness. GAIA~\cite{mialon2023gaia} focuses on complex assistant tasks but lacks explicit support for visual reasoning or generation. The $\tau$-bench~\cite{yao2024tau} framework evaluates tool-agent-user interaction and exposes consistency failures across repeated trials, an issue particularly relevant for iterative analysis workflows. WebArena~\cite{zhou2024webarena} similarly evaluates autonomous agents in realistic web environments, showing significant gaps between benchmark performance and practical reliability.

Tool-centric and multi-agent evaluations further expand the design space. ToolLLM~\cite{qin2023toolllm} evaluates agents across thousands of APIs and provides infrastructure for integrating visualization tools, but it currently lacks visualization-specific metrics. MultiAgentBench~\cite{wu2025multiagentbench} studies collaboration and competition among multiple agents. Holistic evaluation frameworks~\cite{bommasani2023holistic} and agent architecture surveys~\cite{zhao2025ai} provide high-level guidance but require substantial domain-specific adaptation for visualization-centric systems.

\textbf{Multimodal and system-level evaluation.}
Several benchmarks extend agent evaluation to multimodal and system-level settings. VisualWebArena~\cite{koh2024visualwebarena} and MMMU~\cite{yue2024mmmu} assess multimodal understanding and reasoning, highlighting the difficulty of expert-level visual interpretation even for state-of-the-art models. Windows Agent Arena~\cite{bonatti2024windows} emphasizes evaluating agents in their native operational environments, a principle directly relevant to SciVis tools.
Recent work expands agent evaluation beyond leaderboard performance to consider deployment and assessment validity. LLM Evaluate~\cite{saini2025llm} highlights practical concerns such as reliability, scalability, and user-facing quality, while surveys of agent architectures~\cite{zhao2025ai} identify reproducibility and consistency as persistent challenges across agent systems.

At a broader methodological level, Yehudai et al.\ \cite{yehudai2025survey} synthesized open problems in LLM-based agent evaluation, emphasizing reproducibility, generalization, and task diversity. Zhu et al.\ \cite{zhu2025establishing} showed that many existing agentic benchmarks suffer from validity issues in task design and outcome evaluation, and propose best practices for constructing rigorous benchmarks. These challenges become particularly acute in visualization.
Empirical studies~\cite{jana2025evaluating, vazquez2024llms} further demonstrate that LLMs continue to struggle with visualization generation and understanding, underscoring the need for domain-specific, carefully constructed evaluation frameworks.

\vspace{-0.05in}
\subsection{Agentic Workflows and Visualization Systems}

LLMs are embedded in visualization workflows, necessitating evaluation not only of their capabilities but also of their interactions with tools and analysts. Dhanoa et al.\ \cite{dhanoa2025agentic} framed this shift as agentic visualization, defining design patterns that balance autonomous actions and analyst control. Systems in this space span a wide range of integration strategies and levels of agent responsibility.

Recent systems integrate LLMs directly into visualization and analysis pipelines. VOICE~\cite{jia2025voice} provides a multimodal conversational assistant that interprets visualizations and explains features, illustrating how agents can support analytic reasoning. IntuiTF~\cite{wang2025intuitf} shows that MLLMs can steer transfer functions and low-level rendering parameters through iterative visual feedback. CoDA~\cite{chen2025coda} treats visualization more broadly as a collaborative multi-agent workflow in which specialized agents negotiate specification, refinement, and validation.
These ideas extend to tool-integrated systems that operate on scientific data and utilize visualization software. AVA~\cite{liu2024ava} leverages multimodal perception to refine renderings in response to high-level natural language goals. ChatVis~\cite{peterka2025chatvis} translates natural language into ParaView scripts to support non-experts. ParaView-MCP~\cite{liu2025paraview} generalizes this to a persistent tool-using agent capable of multi-step manipulation and stateful interaction across visualization sessions.

Several systems explore self-refinement or distributed coordination beyond single-agent autonomy. VizGenie~\cite{biswas2025vizgenie} dynamically generates and validates visualization scripts, though it largely sidelines human participation. NLI4VolVis~\cite{ai2025nli4volvis} coordinates multiple agents to enable open-vocabulary 3D editing through editable Gaussian splatting~\cite{Tang-PVIS25}, and TexGS-VolVis~\cite{tang2025texgs} introduces textured-Gaussian-splatting for expressive volumetric scene manipulation. Zhang et al.\ \cite{zhang2025automatic} advanced semantic mapping for flow visualization, and InferA~\cite{tam2025infera} implements a multi-agent system tailored to cosmological ensemble exploration. \hot{Latest systems push SciVis agents toward more autonomous scientific workflows, including SASAV~\cite{sun2026sasav} for self-directed scientific analysis and visualization and AI VIS co-scientist~\cite{miao2026toward} for end-to-end generation of task-driven visual analysis applications. 
Ai et al.\ \cite{ai2026hilsva} studied the role of humans in agentic AI systems. 
Vonderhorst et al.\ \cite{vonderhorst2026exploring} further compared how different agent designs and interaction modalities affect multi-step SciVis task performance.
}

While these systems demonstrate increasingly capable agentic behaviors in SciVis, they also pose open methodological challenges regarding reliability, repeatability, and systematic comparison. These challenges motivate SciVisAgentBench, which aims to provide a principled framework for assessing visualization agents operating over scientific data, tools, and workflows.

\vspace{-0.05in}
\subsection{Visualization Benchmarks and Evaluation Methods}

The visualization community has begun developing evaluation frameworks that address the unique challenges of visual representation and analysis. VisEval~\cite{chen2024viseval} benchmarks LLMs on chart readability and generation, revealing substantial failures even for state-of-the-art models. Drawing Pandas~\cite{galimzyanov2025drawing} focuses on plotting code generation and exposes systematic issues in executability and correctness. Beyond surface-level accuracy, Wu et al.\ \cite{wu2023rational} introduced a principled comparison against hypothetical rational agents, offering insight into the actual contribution of visualization to analytical outcomes.

Several systems combine agent design with explicit evaluation protocols. MatPlotAgent~\cite{yang2024matplotagent} argues that agentic visualization requires assessment beyond generic code metrics, particularly to capture iterative refinement. LIDA~\cite{dibia2023lida} introduces visualization-specific metrics such as {\em visualization error rate} and {\em self-evaluated visualization quality}, accounting for both correctness and perceptual quality. PlotGen~\cite{goswami2025plotgen} and nvAgent~\cite{ouyang2025nvagent} extend this line of work by evaluating collaborative, multi-agent visualization workflows that emphasize alignment between user intent and visual output.
Complementary work examines foundational visualization capabilities of language models. Studies on visualization literacy and comprehension~\cite{hong2025llms, bendeck2025llms, islam2024chart} show that LLMs struggle with visual reasoning tasks that are trivial for humans. MLLMs, such as ChartLlama~\cite{chen2023chartllama} and alternative paradigms, including neural machine translation for visualization~\cite{luo2022neural}, address parts of this gap but remain primarily focused on charts rather than full SciVis pipelines.

Related benchmarks from adjacent domains provide practical context but limited coverage. For instance, NL2VIS benchmarks, adapted from NL2SQL~\cite{luo2021synthesizing}, and data-science-oriented agent benchmarks, such as DA-Code~\cite{huang2024dacode}, emphasize query translation and code generation, while ThinkGeo~\cite{shabbir2025thinkgeo} targets tool-augmented agents for remote sensing. These efforts acknowledge exploratory analysis, yet stop short of evaluating end-to-end SciVis workflows involving complex data, rendering pipelines, and interaction. Recently, Ai et al.\ \cite{Ai-GenAI25} called for systematic, reproducible evaluation frameworks for SciVis agents. SVLAT~\cite{do2026svlat} introduces a validated instrument for assessing human SciVis literacy, and a followup work tests MLLM's SciVis literacy~\cite{Do-VISSP26}.
\hot{NL2SciVis~\cite{mathai2026nl2scivis} presents a ParaView benchmark for atomic SciVis operations, but does not cover multi-step workflows across tools and domains.}

Collectively, existing benchmarks and methods demonstrate early progress but remain fragmented, underscoring the need for a comprehensive evaluation framework to assess visualization agents operating on scientific data, tools, and multi-step workflows.

\vspace{-0.075in}
\section{Benchmark Goals and Design Requirements}

SciVisAgentBench is designed to treat evaluation as a central component of SciVis agent design rather than an after-the-fact assessment of completed systems. We note that the lack of systematic, reproducible benchmarks fundamentally constrains progress in SciVis agents. Accordingly, evaluation considerations directly inform task design, data selection, and metric choice in SciVisAgentBench. Guided by this principle, we organize the benchmark around four core design goals.

\textbf{Real-world fidelity.}
To faithfully reflect real-world SciVis practice, the benchmark emphasizes fidelity to the real world. Tasks are grounded in SciVis tools, pipelines, and data formats, rather than abstracted or synthetic proxies. Agents operate over authentic visualization environments (e.g., ParaView~\cite{Ahrens2005ParaView}, napari~\cite{napari2019}, VMD~\cite{VMD1996}), which support interactive data analysis and visualization workflows. 
\hot{However, to enable reproducible and scalable evaluation, the current benchmark focuses on tasks with explicit, verifiable outcomes, excluding open-ended exploration and multiple valid solutions.}

\textbf{Multidimensional evaluation.}
We consider three dimensions: {\em outcome quality}, {\em process behavior}, and {\em efficiency} for benchmark assessment. Outcome correctness is the primary evaluation axis in the current release, while process-level analysis is left to future work due to reproducibility and stability challenges in trajectory assessment. 
\hot{Efficiency is quantified through agent-level execution cost, including time and token usage. Separately, benchmark-level computational tractability is treated as a design requirement to ensure that evaluations can be run repeatedly without excessive overhead.}

\textbf{Reproducibility and determinism.}
Reproducibility is a central design requirement. Each task is anchored by an explicit visualization outcome that serves as the reference for evaluation. When supported by the underlying toolchain, additional artifacts, such as visualization states or scripts, may be recorded to facilitate reproducibility. This enables deterministic re-runs, systematic comparisons across agents, and post hoc inspection of failures. Where full determinism is infeasible due to stochastic model behavior, repeated trials and consistency measures are incorporated to quantify variability and robustness.

\textbf{Extensibility and growth.}
SciVisAgentBench is designed as a living benchmark. Its modular task specifications, evaluator interfaces, and agent abstractions enable the addition of new datasets, tasks, tools, and evaluation metrics without restructuring the benchmark, and facilitate straightforward evaluation of newly developed agents. 

Extensibility is enabled by an established contribution pipeline supported by a public project website 
and structured submission templates. These resources guide domain experts in developing cases across diverse areas, including volume, flow, molecular, bioimage, and topological data analysis. Expert-contributed cases are hosted through a public Hugging Face repository 
to facilitate versioning, access, and reproducibility. In parallel, we actively coordinate with collaborators across national laboratories and universities through recurring meetings that support case development and benchmark expansion.

Contributions are not restricted to tasks solvable by current agents. The benchmark explicitly welcomes forward-looking cases that are well-defined and achievable by human experts, even when they exceed the capabilities of existing systems. By incorporating such tasks, the benchmark functions not only serve as an evaluation tool but also as a means of surfacing open challenges and guiding future research on SciVis agents.

\vspace{-0.075in}
\section{Overview of SciVisAgentBench}
\label{sec:overview}
\hot{
The taxonomy of SciVisAgentBench is derived from both established SciVis principles and the practical requirements of evaluating autonomous SciVis agents. SciVis can be viewed as a process that transforms scientific data into visual representations that support knowledge discovery and sensemaking rather than mere presentation~\cite {telea2014data,wright2007introduction}. Consequently, a benchmark taxonomy should characterize not only the data being visualized, but also the analytical operations performed, the workflow complexity involved, and the scientific contexts in which these activities occur.
Accordingly, we organize benchmark cases along four complementary dimensions: {\em application domain}, {\em data type}, {\em complexity level}, and {\em visualization operation}. 
The taxonomy was iteratively refined during benchmark construction through collaboration with visualization researchers and domain experts across multiple scientific disciplines. Benchmark cases were collected, reviewed, and annotated according to these dimensions, and categories that consistently appeared across expert-designed workflows were retained in the final taxonomy. This process ensures that the taxonomy reflects both established SciVis theory and the practical structure of contemporary SciVis tasks.}

\begin{figure*}[htb]
	\centering
	\includegraphics[width=\linewidth]{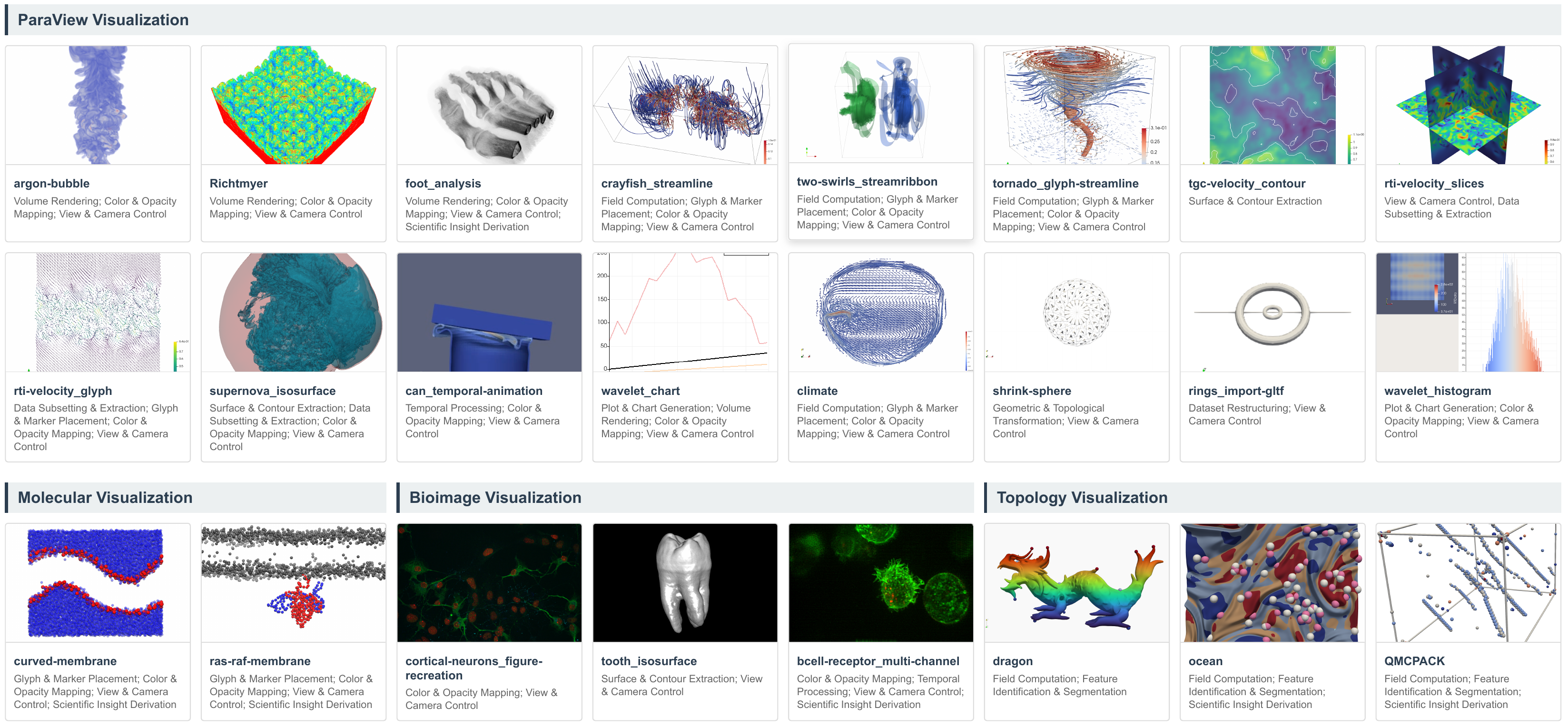}
	\vspace{-0.25in}
    \caption{This gallery displays 24 representative benchmark cases from SciVisAgentBench, organized into four task suites: ParaView visualization, molecular visualization, bioimage visualization, and topology visualization. Each case shows the ground-truth visualization and its associated visualization operations.}
	\label{fig:examples}
	\vspace{-0.1in}
\end{figure*}


{\bf Application domains.}
The benchmark spans a diverse set of scientific domains to reflect the breadth of real-world visualization practice. Current benchmark cases draw from astronomy, medical science, biology, physics, earth system science, mathematics, and chemistry.

{\bf Data types.}
Datasets are selected to represent major field structures in SciVis, including scalar, vector, and tensor fields, as well as multivariate and time-varying data, which fundamentally influence visualization design and analysis strategies~\cite{tory2004vistaxonomy,telea2014data}. To keep evaluation costs practical, some datasets are downsampled in spatial resolution or temporal duration while preserving their analytical characteristics. This design supports realistic tasks without excessive resource requirements.

{\bf Complexity levels.}
The complexity of cases is categorized into three tiers: \emph{operations}, \emph{tasks}, or \emph{workflows} based on the highest-level procedural scope. 
An operation is a defined procedure with explicit inputs and outputs.
A task is a structured sequence of operations performed to accomplish a clearly defined goal.
A workflow is a structured, potentially multi-stage process comprising interrelated tasks that collectively achieve a broader objective. 
Only task-level and workflow-level entries are reported as benchmark cases, while operations serve as compositional building blocks within these higher-level tiers.

{\bf Visualization operations.}
\hot{
Each benchmark case is annotated with one or more visualization operations that characterize the primary actions performed. The taxonomy intentionally includes both enabling operators and work operators~\cite{laha2015classification}, reflecting the fact that SciVis workflows require bridging the gap between raw data manipulation and the delivery of scientific insight~\cite{gooch2005illustrative}.}
For the complete list of operations, refer to Appendix~\ref{subsec:app-vot}. 
Selected cases are shown in Figure~\ref{fig:examples}.

\hot{The inclusion of both low-level visualization operations (e.g., Color \& Opacity Mapping, View \& Camera Control) and higher-level analytical operations (e.g., Scientific Insight Derivation) is intentional. Prior empirical studies of SciVis workflows distinguish between enabling operators, which prepare data for inspection, and work operators, which correspond to scientific analysis objectives~\cite{laha2015classification}. Both are essential components of real-world SciVis workflows, as scientific insights emerge from the coordinated use of visualization mechanics and analytical reasoning. By evaluating agents across this spectrum, we measure not only whether agents can correctly manipulate visualization tools, but also whether they can use them to support scientific understanding.}

{\bf Benchmark and test cases.}
The list of all benchmark cases with their corresponding taxonomy annotations is provided in Appendix~\ref{appendix:cases}. For all test cases with task descriptions, evaluation rubrics, ground-truth visualizations, and agent results, please refer to our project page.

{\bf Agent interfaces and evaluated systems.}
SciVisAgentBench is designed to support the evaluation of a wide range of LLM-based SciVis agents that differ in architecture, modality, and interaction style. Rather than prescribing a single-agent design, the benchmark abstracts away agent interfaces to enable consistent and fair evaluation across heterogeneous systems.
First, the benchmark supports \textbf{tool-using agents} that directly manipulate SciVis software through APIs or protocol-based interfaces, such as MCP-style tool invocation. These agents operate over visualization environments and maintain state across multi-step interactions. 
Second, \textbf{code-generating agents} are supported, in which agents produce executable visualization scripts or pipelines (e.g., PvPython scripts) that are subsequently run by external tools to generate visualization outputs. 
Third, the benchmark supports the evaluation of \textbf{human-like interface agents} that interact with visualization software in the same way as human users, using screenshots, mouse clicks, keyboard input, scrolling, and other GUI-level interactions rather than structured APIs or code execution.
In addition, the benchmark accommodates \textbf{multimodal and multi-agent systems} that combine natural language, vision, and tool feedback, including architectures in which multiple agents collaborate or specialize across different subtasks.

Importantly, the benchmark is intentionally agnostic to how an agent achieves a solution. By decoupling evaluation logic from agent implementation details, SciVisAgentBench enables systematic comparison across these diverse paradigms, allowing agents with fundamentally different internal designs to be evaluated against the same tasks, datasets, and evaluation criteria.


\vspace{-0.075in}
\section{Evaluation Framework}

\subsection{Evaluation Taxonomy}

SciVisAgentBench employs a practical evaluation taxonomy that reflects how SciVis agents are utilized and evaluated in real-world workflows. We organize evaluation along three complementary dimensions: \emph{outcome}, \emph{process}, and \emph{efficiency}. Such a taxonomy supports fair comparison across heterogeneous agent architectures while remaining compatible with reproducible, large-scale benchmarking. Together, these three dimensions establish a coherent evaluation framework for SciVis agents that supports consistent comparison across systems and practical deployment at benchmark scale.

\textbf{Outcome-based evaluation.}
%
Outcome-based evaluation examines the relationship between task specifications and visualization outputs, treating the agent as a black box. This approach enables direct comparison across diverse agent designs (such as script-generating agents, tool-using agents, and multimodal multi-agent systems) without requiring access to internal states or reasoning traces. In SciVisAgentBench, evaluation centers on whether the generated visualization satisfies task constraints and semantic requirements. A key challenge is the non-uniqueness of valid solutions: different visualizations may convey the same scientific insight, complicating automated scoring. To mitigate this, one can either increase task constraints to narrow the solution space or focus on tightly scoped tasks with no branching outcomes. 
\hot{In the current version, we adopt the latter strategy: we include only tasks with a single, explicit visualization result and exclude cases with multiple equally valid outputs. To enforce this constraint, benchmark tasks are first authored by domain experts with detailed visualization requirements, including rendering methods, camera viewpoints, color mappings, and, when applicable, lighting conditions. The task is then independently reviewed and executed by visualization researchers to identify potential branching outcomes. Tasks that admit substantially different yet equally reasonable solutions are either refined with additional constraints or excluded from the benchmark.}
This design ensures reproducibility, robustness, and fair comparison across heterogeneous agent architectures. \hot{However, it does not capture open-ended SciVis tasks that allow multiple valid visualization outcomes. We view this as a pragmatic tradeoff to enable reliable, scalable evaluation, while support for open-ended tasks remains future work.}

\textbf{Process-based evaluation.}
Process-based evaluation examines how an agent reaches a solution, including its action sequence, tool usage, and intermediate decisions. It is useful for diagnosing failures, assessing generalization, and distinguishing systematic reasoning from trial-and-error behavior. Key dimensions include task complexity, tool choice and coordination, and procedural efficiency.

Although conceptually central for understanding SciVis agent behavior, process-based evaluation is not fully realized in the current SciVisAgentBench. The main challenge is establishing reliable ground-truth action trajectories for complex tasks. Manual reference paths are labor-intensive, and automated alternatives are limited. Recent MCP-based evaluations have explored using successful agent trajectories as implicit ground truth~\cite{liu2025mcpeval}, constrained-prompt executions as references~\cite{yin2025livemcp}, or LLM-based evaluators to directly judge trajectories~\cite{wang2025mcp}. These methods assume a largely unique or outcome-insensitive action sequence. This assumption rarely holds in SciVis: multiple sequences can produce the same visualization, while early tool or parameter choices can dramatically affect results.


\hot{Consequently, trajectory-level evaluation is currently impractical for complex, path-dependent SciVis tasks. Nevertheless, SciVisAgentBench already records execution trajectories, tool interactions, intermediate artifacts, and outcomes for all benchmark runs. Combined with the benchmark's containerized execution environment, these records provide a foundation for future process-based evaluation, trajectory analysis, and agent training. In particular, successful trajectories and reward signals may serve as valuable supervision for agentic reinforcement learning and post-training methods targeting scientific data analysis and discovery. Developing reliable trajectory evaluators and reward functions remains an important direction for future work.}

\textbf{Efficiency-aware evaluation.}
\hot{Efficiency evaluation focuses on agent-level resource usage, including execution time, interaction steps, and token consumption, reflecting the cost of deploying SciVis agents in scientific workflows. Separately, SciVisAgentBench is designed to remain computationally tractable, avoiding excessive evaluation time, tool invocations, or evaluator overhead. We treat this as a benchmark design consideration rather than an evaluation dimension.}

\begin{figure*}[htb]
	\centering
	\includegraphics[width=0.95\linewidth]{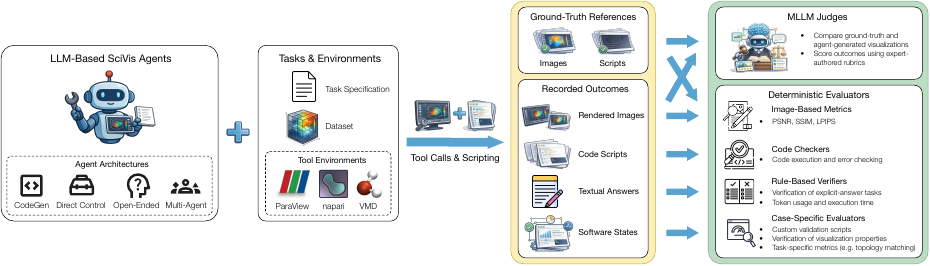}
	\vspace{-0.1in}
    \caption{Overview of the SciVisAgentBench execution and evaluation pipeline. 
    LLM-based SciVis agents interact with SciVis tools and datasets to perform benchmark tasks and produce recorded outcomes. 
    These outcomes are evaluated against ground-truth references using a combination of MLLM judges and deterministic evaluators, yielding visualization quality, efficiency, and final benchmark scores for recorded outcomes.}
	\label{fig:overview}
    \vspace{-0.1in}
\end{figure*}

\vspace{-0.05in}
\subsection{Task Definition and Outcome-Centric Structuring}

We organize evaluation around well-defined, outcome-centric tasks grounded in scientific datasets. Each task is paired with a natural-language SciVis description, authored and refined by visualization experts, that specifies a concrete analysis or visualization goal. To ensure evaluability and reproducibility, all tasks included in the benchmark are carefully curated to admit a single, explicit visualization outcome.

\hot{Benchmark tasks are collected through collaboration with domain experts and visualization researchers across multiple scientific disciplines. Candidate tasks are mostly created by domain experts and then selected by visualization researchers based on four criteria: (1) relevance to real-world SciVis workflows, (2) clear evaluation objectives, (3) feasibility for human experts using existing tools, and (4) alignment with the benchmark taxonomy.}

Each task is anchored by a mandatory ground-truth artifact, typically a reference visualization image. When the underlying toolchain naturally exposes additional artifacts, such as executable scripts or visualization engine states, these are also provided as ground truth. For assessment, every task includes expert-authored evaluation rubrics that describe the required properties of a desirable visualization outcome. These rubrics are used to score outcome quality, while efficiency metrics capture token usage and execution time. By default, outcome correctness dominates the task score, with efficiency treated as a secondary but explicit consideration.

Task specifications and rubrics are serialized in a structured YAML format compatible with existing evaluation frameworks such as promptfoo~\cite{promptfoo}, enabling rapid text-based testing. However, such frameworks cannot directly verify visualization artifacts. They are used only as auxiliary infrastructure rather than as the primary evaluation mechanism.

A single dataset may support multiple tasks of varying scope and difficulty. To accommodate 
\hot{multi-step} scientific workflows without relying on unstable process-based evaluation, complex tasks are decomposed into smaller, verifiable checkpoints with explicit intermediate outcomes. While success on these subtasks does not guarantee end-to-end workflow completion, this decomposition enables reliable outcome-based evaluation within current methodological constraints.

\vspace{-0.05in}
\subsection{Multimodal Outcome Evaluation with LLM Judges}

SciVisAgentBench uses MLLMs as outcome evaluators to assess visualization quality. For each task, the evaluator is provided with the ground-truth visualization, the agent-generated visualization, and a set of expert-authored evaluation rubrics describing the required visual and semantic properties. The LLM judge scores each rubric on a 0-10 scale based on how well the agent output satisfies the specified criteria. 
Prior work has shown that LLM-based judges can be sensitive to prompt formulation and rubric ordering~\cite{li2025evaluating}. To mitigate ordering bias, we adopt a rubric shuffling strategy that randomly permutes the order of evaluation criteria across trials. In addition, MLLM judgments may not always align with human preferences in domains with limited training data~\cite{zheng2023judging, gu2024survey, Wei-arXiv2024}, such as SciVis.
To assess the reliability of this evaluation strategy, we conduct dedicated validation studies, including a human-LLM alignment analysis and a prompt- and presentation-robustness study, which are reported in Section~\ref{subsec:outcome_valid}.
These analyses examine whether high LLM-assigned scores reliably correspond to successful task completion and visually correct outcomes.

\vspace{-0.05in}
\subsection{Deterministic Evaluators}
\label{subsec:deterministic-evaluator}

In addition to LLM-based judges, SciVisAgentBench incorporates rule-based evaluators that provide deterministic, reproducible assessment signals when the task structure and tool support allow. These evaluators complement multimodal judgment by grounding evaluation in executable checks and quantitative measurements.

\textbf{Image-based metrics.}
For tasks whose ground-truth visualizations are rendered under controlled conditions (e.g., fixed camera pose, lighting, and background), the benchmark supports quantitative evaluation of image-based similarity between agent-generated outputs and ground-truth visualizations. We consider pixel-level and perceptual metrics, including PSNR for measuring reconstruction fidelity, SSIM for assessing structural similarity~\cite{wang2004ssim}, and LPIPS for capturing perceptual alignment~\cite{zhang2018lpips}. Higher PSNR and SSIM values indicate closer correspondence to the reference visualization, while lower LPIPS values indicate greater perceptual similarity. To reduce sensitivity to viewpoint, these metrics are computed from multiple canonical viewpoints (front, side, and diagonal) and averaged per task. 

To prevent partial task completion from inflating metric scores, we instead report scaled variants computed from the aggregated scores: 
$\mathrm{PSNR}_{\text{scaled}} = r \times \overline{\mathrm{PSNR}}$,
$\mathrm{SSIM}_{\text{scaled}} = r \times \overline{\mathrm{SSIM}}$, and 
$\mathrm{LPIPS}_{\text{scaled}} = 1 - r \times (1 - \overline{\mathrm{LPIPS}})$,
where $\overline{(\cdot)}$ denotes the metric value averaged over all evaluated cases, $r = N_{\text{pass}}/N_{\text{total}}$, and $N_{\text{pass}}$ and $N_{\text{total}}$ denote the numbers of successfully completed and total cases. These normalizations prevent incomplete evaluations from artificially inflating scores, ensuring fair comparison across agents with different completion rates.

\textbf{Code checkers.}
For tasks completed via code generation (e.g., PvPython scripts), the benchmark provides reference code. Evaluation proceeds by verifying correct script generation and placement, executing the agent-generated code, and checking for successful visualization output. 
We also explore measuring similarity between reference and generated scripts using CodeBERT~\cite{feng2020codebert}. However, in practice, this signal proved unreliable for SciVis workflows: functionally correct and incorrect scripts often received similar similarity scores, and the metric failed to capture nuanced but critical differences. Therefore, we do not incorporate CodeBERT similarity into the final scoring.

\textbf{Rule-based verifiers.}
In addition to visualization-centric tasks, SciVisAgentBench includes tasks with discrete outputs, such as multiple-choice and binary (yes/no) questions. In these cases, evaluation is performed using deterministic, rule-based verifiers that compare the agent's response to ground-truth answers. This form of verification provides an unambiguous signal of correctness by framing analytical reasoning tasks as questions with a single explicit answer.

\textbf{Case-specific evaluators.}
When tasks cannot be reliably assessed using the evaluators above, SciVisAgentBench supports case-specific evaluation scripts. This is particularly important for tasks whose expected outputs are data rather than visualization artifacts. For example, critical point identification may yield a set of spatial coordinates that cannot be meaningfully evaluated solely through image-based comparison. Custom scripts enable deterministic verification of such outputs.

Case-specific evaluators may also verify visualization configurations produced by the agent. For visualization systems that expose explicit internal states (e.g., ParaView), SciVisAgentBench provides ground-truth software states. Case-specific scripts could verify that required visualization properties, such as isovalue thresholds, color mappings, rendering modes, or seed placement, are met.

In addition, the benchmark introduces reusable evaluators for topology visualization tasks, which generalize across multiple cases and reduce the need for task-level customization. For critical point prediction (for scalar fields) and degenerate point prediction (for tensor fields), an optimal pairing is computed between ground truth and predicted points. A similarity score is derived from the pairing. For persistence diagrams, the similarity score is computed from the Wasserstein distance \cite[page 183]{edelsbrunner2022computational}. For merge trees, the evaluation score is based on the Wasserstein distance of their persistence diagrams as well as the partial fused Gromov Wasserstein distance as adapted to merge trees by Li et al.~\cite{li2025flexible}. For Morse-Smale segmentation, the Hungarian algorithm is used to compute region pairings that maximize the dice score; the similarity score is based on the dice score. For asymmetric tensor fields, the eigenvector and eigenvalue partitions \cite{zhang2008asymmetric} are scored using the dice score derived from the region labels provided by the partitions.

\vspace{-0.05in}
\subsection{Evaluation Pipeline}

As shown in Figure~\ref{fig:overview}, SciVisAgentBench provides an end-to-end evaluation pipeline that enables rapid and reproducible assessment of SciVis agents with minimal integration effort. Agents must conform to a lightweight interface by inheriting a predefined base class and implementing a single method that encapsulates the task execution logic. This design enables the evaluation of agents with diverse internal architectures and tool usage strategies within a unified framework.

During evaluation, the framework initializes the execution environment, loads datasets and task specifications, and sequentially invokes the agent on each test case. Agents interact with the provided tools or their own integrated toolchains and must store outcome artifacts, such as rendered images, generated scripts, or software states, at predefined locations. After task execution, the framework verifies the presence of expected outputs and performs post-processing when necessary, such as re-rendering visualizations under standardized camera and background configurations.

Evaluation is then conducted using a combination of MLLM-based judges and rule-based evaluators, depending on task type. In parallel, the framework records execution time, token usage, and estimated cost. Results from individual tasks are aggregated into per-task reports, and final scores are summarized across the selected task suite. These aggregated results are used to populate an online leaderboard, enabling systematic comparison of agents and their underlying LLM backbones.
	
\vspace{-0.075in}
\section{Validity Study}

Following Zhu et al.~\cite{zhu2025establishing}, we examine two key aspects of benchmark validity: \emph{task validity} and \emph{outcome validity}. Task validity requires that a benchmark task be solvable if and only if an agent possesses the intended capability, while outcome validity requires that the evaluation signal faithfully reflects true task success. In this section, we analyze our task design to support task validity and conduct human-LLM alignment and robustness studies to assess the validity of the outcomes.

\vspace{-0.05in}
\subsection{Task Validity}

\hot{Our task validity argument relies on two complementary properties. First, the benchmark taxonomy is designed to represent the core capabilities required by SciVis agents. Consequently, successful completion of benchmark tasks requires agents to demonstrate at least one of these intended capabilities. Second, benchmark cases are explicitly annotated according to the taxonomy, ensuring that every task can be mapped to the capability it is intended to evaluate. Together, the taxonomy-capability correspondence provides a direct link between benchmark performance and the target capabilities of SciVis agents.}

Task validity requires that the target capability be tightly coupled with task success. To support this property, SciVisAgentBench is grounded in a structured taxonomy that covers application domains, data types, complexity levels, and visualization operations (see Section~\ref{sec:overview} and Appendix~\ref{appendix:taxonomy}). 

The current benchmark cases achieve broad, though intentionally non-uniform, coverage of the taxonomy. This distribution reflects both the inherent frequency of different SciVis scenarios and the expert-driven nature of case construction. For example, tensor-field datasets are relatively rare in practical workflows and therefore appear less frequently. Similarly, geometric and topological transformation operations are less common in typical analytic pipelines.

We also deliberately exclude certain categories from the current release. 
Data sampling and resolution control operations, while important, are often treated as preprocessing steps, typically require large-scale datasets for meaningful evaluation, and can be substantially influenced by external algorithms (e.g., implicit neural representations). Including them would increase computational overhead without directly probing agentic reasoning capabilities. Our design prioritizes tasks that most clearly reflect autonomous agentic workflows, while keeping the benchmark computationally tractable.

\begin{table}[t]
\centering
\small
\caption{Inter-rater reliability within human experts and LLM judges (Gemini-3.1-Pro, Claude-Opus-4.6, GPT-5.2). Here, $n$ denotes the number of individual judgments, where each judgment is a full evaluation of all cases by a single expert or a single model run. The best values are highlighted in bold.
}
\label{tab:reliability_comparison}
\vspace{-0.1in}
\begin{tabular}{lcc}
\toprule
Evaluator & Krippendorff's $\alpha$ $\uparrow$ & ICC(2,1) $\uparrow$ \\
\midrule
Human Experts ($n{=}12$) & 0.669 & 0.673 \\
Human Experts (Outliers Removed, $n{=}10$) & 0.719 & 0.723 \\
LLM Judges (Cross-Model Aggregate, $n{=}15$) & \textbf{0.817} & \textbf{0.819} \\
\bottomrule
\end{tabular}
	\vspace{-0.1in}
\end{table}

\begin{table}[htb]
\centering
\small
\caption{Alignment between LLM judges and human expert evaluations (21 cases, 65 ratings). 
}
\label{tab:human_llm_alignment}
\vspace{-0.1in}
\begin{tabular}{lcccc}
\toprule
Model & Pearson $r$ $\uparrow$ & Spearman $\rho$ $\uparrow$ & MAE $\downarrow$ & RMSE $\downarrow$ \\
\midrule
Gemini-3.1-Pro & \textbf{0.808} & \textbf{0.830} & 1.537 & 2.138 \\
Claude-Opus-4.6 & 0.806 & 0.815 & \textbf{1.469} & \textbf{1.945} \\
GPT-5.2 & 0.764 & 0.789 & 1.691 & 2.393 \\
\bottomrule
\end{tabular}
	\vspace{-0.1in}
\end{table}

\begin{table}[htb]
\centering
\small
\caption{Stability of LLM judges across five repeated trials. 
}
\label{tab:judge_stability}
\vspace{-0.1in}
\begin{tabular}{lccc}
\toprule
Model & Stability Score $\uparrow$ & Mean Std $\downarrow$ & Median Std $\downarrow$ \\
\midrule
Gemini-3.1-Pro & 0.929 & 0.778 & 0.548 \\
Claude-Opus-4.6 & \textbf{0.975} & \textbf{0.271} & \textbf{0.000} \\
GPT-5.2 & 0.960 & 0.441 & 0.447 \\
\bottomrule
\end{tabular}
	\vspace{-0.1in}
\end{table}

\vspace{-0.05in}
\subsection{Outcome Validity}
\label{subsec:outcome_valid}

Outcome validity requires that successful task completion corresponds to a positive evaluation signal. In SciVisAgentBench, this primarily concerns the reliability of MLLM judges for assessing visualization quality. We examine two properties of the LLM judges:
\begin{myitemize}
\vspace{-0.05in}
\item \textbf{Accuracy and agreement}:\ The judge should demonstrate strong accuracy, self-consistency, and agreement with human experts.
\item \textbf{Robustness}:\ The judge should remain stable under prompt and presentation variations.
\vspace{-0.05in}
\end{myitemize}

\textbf{Human-LLM alignment study.}
As shown in Figure~\ref{fig:examples}, we constructed a stratified sample of 24 benchmark cases to ensure coverage across application domains, data types, complexity levels, and visualization operations. Among these, 21 cases (excluding the three cases from topology visualization) require vision-based evaluation and were therefore included in the human-LLM alignment study. For each selected case, we generated agent outputs using Claude-Sonnet-4.5 as the backbone model.

\hot{We developed an online evaluation platform and recruited 12 SciVis experts to independently score the results. The study was approved under the University of Notre Dame's IRB protocol (No.\ 18-01-4334), and informed consent was obtained from all experts.} They were provided with the same information as the LLM judge, including the task description, ground-truth visualization, agent output image, evaluation rubrics, and scoring guidelines. Each rubric was scored on a 0-10 scale, yielding 65 ratings across the 21 cases.

As shown in Table~\ref{tab:reliability_comparison}, we assessed inter-rater reliability within human experts and LLM-based evaluators using Krippendorff's $\alpha$ and the intraclass correlation coefficient (ICC). For both metrics, higher values correspond to stronger inter-rater reliability and greater agreement among judgments. The 12 human experts exhibit substantial agreement. We further conducted a leave-one-out reliability analysis to measure the influence of individual experts on overall agreement. This analysis reveals two experts whose ratings consistently deviate from the rest of the group: removing either expert leads to a noticeable increase in agreement, shifting the reliability level from insufficient to tentative conclusions under the standard interpretation of Krippendorff's $\alpha$.
Following this data-driven criterion, we excluded the two outliers and recomputed the reliability statistics using the remaining 10 experts. Inter-rater agreement improves after outlier removal, and the filtered human evaluations serve as the reference signal for subsequent human-LLM alignment analysis.

We further assessed the consistency of multiple LLM judges (Gemini-3.1-Pro, Claude-Opus-4.6, and GPT-5.2). For each model, we ran 5 independent evaluation runs, yielding 15 individual judgments. Overall, different LLM judges exhibit higher inter-rater agreement than human experts, indicating strong consistency in their evaluation behavior. \hot{We note that lower agreement among human experts may also reflect legitimate differences in scientific background, experience, and subjective preferences rather than inferior evaluation quality.}
We then evaluated how well LLM judgments align with human evaluations by computing Pearson correlation, Spearman rank correlation, mean absolute error (MAE), and root mean square error (RMSE) between human and model judgments. Human ratings were averaged across experts, and model ratings were averaged across repeated runs.

Table~\ref{tab:human_llm_alignment} summarizes the alignment results. 
\hot{All three models exhibit strong positive correlation with human judgments, suggesting that LLM-based judges can approximate expert evaluations within the evaluated benchmark setting.
However, these results do not indicate equivalence between LLM and human judgment, particularly for open-ended visualization tasks that involve substantial subjective interpretation.} Gemini-3.1-Pro achieves the highest correlation with human scores, while Claude-Opus-4.6 yields the lowest RMSE and the closest score distribution to human ratings.

\textbf{Prompt and presentation robustness.}
To assess robustness, we evaluated the LLM judge's sensitivity to controlled perturbations in prompts and presentation. Specifically, we varied rubric phrasing, scoring scale descriptions, the order of image presentation, and the visibility of ground-truth images. In addition, each evaluation was repeated five times to measure stochastic variability in model responses.

We quantified robustness using a \emph{judge stability} metric that measures score consistency across repeated trials and perturbations. For each case, we computed the normalized standard deviation of the judge scores across all experimental conditions and aggregated the results across cases. Formally, the stability score is defined as
$S_{\text{stable}} = 1 - (\sum_{i=1}^{N} \sigma_i/R)/N$,
where $\sigma_i$ is the standard deviation of the judge scores for case $i$ across perturbations, $R=11$ is the scoring range (0–10), and $N$ is the number of evaluated cases. Higher values indicate more consistent judgments and, therefore, greater robustness of the evaluation signal.
Table~\ref{tab:judge_stability} reports stability across models. All LLM judges show high stability ($S_{\text{stable}}>0.92$), with Claude-Opus-4.6 achieving the highest score.


\hot{
Overall, MLLM judges provide reliable and stable evaluation signals. Human-LLM correlations approach human inter-rater agreement, supporting the use of LLM-based judges in SciVisAgentBench. We note that higher agreement is not necessarily better, as lower agreement among human experts may reflect diverse assessments. Claude-Opus-4.6 is selected not because it is objectively superior, but because it exhibits the strongest alignment with the 12 SciVis experts while maintaining high stability.
We further observe that alignment is strongest when evaluation criteria are explicit and visually grounded (e.g., whether a rendered structure has the correct color or opacity). In contrast, discrepancies are more likely to arise when subjective judgment or higher-level interpretation is required, such as assessing overall visualization quality or performing multi-step reasoning over visual evidence. Therefore, LLM judges should be viewed as scalable approximations of expert evaluation rather than replacements for human assessment.}

\begin{table*}[t]
\centering
\small
\caption{Benchmark performance across all five task suites of SciVisAgentBench using {\bf Claude-Opus-4.6} as the LLM judge for evaluation. For each agent+model setting, we report overall score, completion rate, pass@${\{1,2,3\}}$ (i.e., success in at least one of the first $k$ trials), and pass$\textasciicircum{\{1,2,3\}}$ (i.e., success in all $k$ trials). Scores and completion rates are reported as mean$\pm$std across three repeated trials. 
}
\label{tab:main_results_all_parts_opus46}
\vspace{-0.1in}
\begin{adjustbox}{width=\textwidth}
\begin{tabular}{clcccccccc}
\toprule
Task Suite & Setting & Overall Score $\uparrow$ & Completion Rate $\uparrow$ & pass@1 $\uparrow$ & pass@2 $\uparrow$ & pass@3 $\uparrow$ & pass$\textasciicircum1$ $\uparrow$ & pass$\textasciicircum2$ $\uparrow$ & pass$\textasciicircum3$ $\uparrow$ \\
\midrule

\multirow{6}{*}{\shortstack{ParaView\\Visualization}}
& ChatVis+GPT-5.2 & 30.77$\pm$1.10 & 44.44$\pm$1.20 & 0.382 & 0.528 & 0.604 & 0.382 & 0.236 & 0.167 \\
& ChatVis+Claude-Sonnet-4.5 & 37.37$\pm$3.02 & 54.17$\pm$3.61 & 0.458 & 0.542 & 0.562 & 0.458 & 0.375 & 0.312 \\
& ParaView-MCP+GPT-5.2 & 24.63$\pm$3.72 & 46.53$\pm$6.36 & 0.236 & 0.271 & 0.292 & 0.236 & 0.201 & 0.188 \\
& ParaView-MCP+Claude-Sonnet-4.5 & 26.43$\pm$8.80 & 53.47$\pm$15.07 & 0.257 & 0.347 & 0.417 & 0.257 & 0.167 & 0.146 \\
& Claude-Code+Claude-Sonnet-4.5 & \textbf{62.57$\pm$0.51} & \textbf{97.92$\pm$2.08} & 0.736 & 0.875 & 0.917 & 0.736 & 0.597 & 0.500 \\
& Codex+GPT-5.2 & 60.17$\pm$1.43 & 95.14$\pm$2.41 & \textbf{0.771} & \textbf{0.910} & \textbf{0.938} & \textbf{0.771} & \textbf{0.632} & \textbf{0.521} \\

\midrule

\multirow{4}{*}{\shortstack{Molecular\\Visualization}}
& GMX-VMD-MCP+GPT-5.2 & 45.67$\pm$7.48 & 66.67$\pm$11.75 & 0.564 & 0.692 & 0.769 & 0.564 & 0.436 & 0.385 \\
& GMX-VMD-MCP+Claude-Sonnet-4.5 & 60.23$\pm$4.39 & \textbf{97.44$\pm$4.44} & 0.846 & 0.846 & 0.846 & 0.846 & 0.846 & \textbf{0.846} \\
& Claude-Code+Claude-Sonnet-4.5 & 61.47$\pm$6.78 & 94.87$\pm$4.44 & \textbf{0.897} & 0.923 & 0.923 & \textbf{0.897} & \textbf{0.872} & \textbf{0.846} \\
& Codex+GPT-5.2 & \textbf{62.30$\pm$6.32} & 94.87$\pm$8.88 & 0.872 & \textbf{0.949} & \textbf{1.000} & 0.872 & 0.795 & 0.769 \\

\midrule

\multirow{4}{*}{\shortstack{Bioimage\\Visualization}}
& BioImage-Agent+GPT-5.2 & \textbf{66.67$\pm$4.28} & 81.82$\pm$0.00 & \textbf{0.788} & \textbf{0.818} & \textbf{0.818} & \textbf{0.788} & \textbf{0.758} & \textbf{0.727} \\
& BioImage-Agent+Claude-Sonnet-4.5 & 57.67$\pm$6.39 & 78.79$\pm$5.25 & 0.636 & 0.697 & 0.727 & 0.636 & 0.576 & 0.545 \\
& Claude-Code+Claude-Sonnet-4.5 & 52.83$\pm$9.80 & \textbf{90.91$\pm$9.09} & 0.697 & 0.758 & \textbf{0.818} & 0.697 & 0.636 & 0.636 \\
& Codex+GPT-5.2 & 41.90$\pm$4.69 & 75.76$\pm$10.50 & 0.576 & 0.697 & 0.727 & 0.576 & 0.455 & 0.364 \\

\midrule

\multirow{4}{*}{\shortstack{Topology\\Visualization}}
& TopoPilot+GPT-5.2 & 31.13$\pm$2.71 & 55.56$\pm$0.00 & 0.148 & 0.185 & 0.222 & 0.148 & 0.111 & 0.111 \\
& TopoPilot+Claude-Sonnet-4.5 & 32.07$\pm$1.01 & 55.56$\pm$0.00 & 0.185 & 0.296 & 0.333 & 0.185 & 0.074 & 0.000 \\
& Claude-Code+Claude-Sonnet-4.5 & 45.23$\pm$8.81 & 59.26$\pm$6.42 & 0.444 & 0.519 & 0.556 & 0.444 & 0.370 & 0.333 \\
& Codex+GPT-5.2 & \textbf{76.43$\pm$10.06} & \textbf{85.19$\pm$6.42} & \textbf{0.778} & \textbf{0.889} & \textbf{0.889} & \textbf{0.778} & \textbf{0.667} & \textbf{0.556} \\

\midrule

\multirow{4}{*}{\shortstack{Object\\Identification}}
& ParaView-MCP+GPT-5.2 & 26.73$\pm$7.36 & 49.38$\pm$11.32 & 0.358 & \textbf{0.593} & \textbf{0.741} & 0.358 & 0.123 & 0.037 \\
& ParaView-MCP+Claude-Sonnet-4.5 & 42.17$\pm$0.50 & \textbf{92.59$\pm$0.00} & 0.185 & 0.284 & 0.333 & 0.185 & 0.086 & 0.037 \\
& Claude-Code+Claude-Sonnet-4.5 & 41.50$\pm$3.55 & 83.95$\pm$9.32 & 0.358 & 0.556 & 0.704 & 0.358 & 0.160 & 0.111 \\
& Codex+GPT-5.2 & \textbf{43.33$\pm$5.28} & 92.59$\pm$9.80 & \textbf{0.395} & 0.519 & 0.556 & \textbf{0.395} & \textbf{0.272} & \textbf{0.185} \\

\bottomrule
\end{tabular}
\end{adjustbox}
	\vspace{-0.1in}
\end{table*}

\begin{table}[t]
\centering
\small
\caption{Image-based evaluation metrics on ParaView visualization tasks. Values are reported as mean$\pm$std across three repeated trials. 
}
\label{tab:paraview_image_metrics}
\vspace{-0.1in}
\begin{adjustbox}{width=\columnwidth}
\begin{tabular}{lccc}
\toprule
Setting & $\mathrm{PSNR}_{\text{scaled}} \uparrow$ & $\mathrm{SSIM}_{\text{scaled}} \uparrow$ & $\mathrm{LPIPS}_{\text{scaled}} \downarrow$ \\
\midrule
ChatVis+GPT-5.2 & 9.63$\pm$0.67 & 0.44$\pm$0.01 & 0.57$\pm$0.01 \\
ChatVis+Claude-Sonnet-4.5 & 10.50$\pm$0.97 & 0.50$\pm$0.04 & 0.50$\pm$0.04 \\
ParaView-MCP+GPT-5.2 & 9.36$\pm$1.27 & 0.46$\pm$0.06 & 0.57$\pm$0.05 \\
ParaView-MCP+Claude-Sonnet-4.5 & 12.00$\pm$2.96 & 0.54$\pm$0.14 & 0.49$\pm$0.14 \\
Claude-Code+Claude-Sonnet-4.5 & 20.99$\pm$0.68 & \textbf{0.92$\pm$0.02} & \textbf{0.10$\pm$0.02} \\
Codex+GPT-5.2 & \textbf{21.27$\pm$1.02} & \textbf{0.92$\pm$0.02} & 0.10$\pm$0.03 \\
\bottomrule
\end{tabular}
\end{adjustbox}
	\vspace{-0.1in}
\end{table}

\begin{table}[t]
\centering
\small
\caption{Token usage across all five task suites of SciVisAgentBench. Input and output tokens are reported as mean$\pm$std across three repeated trials. Token counts are shown using K (thousands) and M (millions) for readability. Cached tokens are counted as regular input tokens for consistent accounting and comparison across settings. 
}
\label{tab:token_cost_all_parts}
\vspace{-0.1in}
\begin{adjustbox}{width=\columnwidth}
\begin{tabular}{clcc}
\toprule
Task Suite & Setting & Input Tokens $\downarrow$ & Output Tokens $\downarrow$\\
\midrule
\multirow{6}{*}{\shortstack{ParaView\\Visualization}}
& ChatVis+GPT-5.2          & 156.64K$\pm$8.02K & 180.56K$\pm$8.84K \\
& ChatVis+Claude-Sonnet-4.5 & {\bf 116.83K$\pm$7.59K} & {\bf 152.70K$\pm$6.88K} \\
& ParaView-MCP+GPT-5.2     & 5.71M$\pm$0.61M   & 33.30K$\pm$1.99K  \\
& ParaView-MCP+Claude-Sonnet-4.5 & 28.51M$\pm$3.30M  & 380.30K$\pm$42.53K \\
& Claude-Code+Claude-Sonnet-4.5 & 39.49M$\pm$6.62M & 425.32K$\pm$55.52K \\
& Codex+GPT-5.2 & 45.57M$\pm$9.47M & 396.60K$\pm$23.27K \\
\midrule
\multirow{4}{*}{\shortstack{Molecular\\Visualization}}
& GMX-VMD-MCP+GPT-5.2      & {\bf 1.56M$\pm$0.17M} & {\bf 28.27K$\pm$6.57K} \\
& GMX-VMD-MCP+Claude-Sonnet-4.5   & 5.89M$\pm$1.00M & 82.90K$\pm$12.53K \\
& Claude-Code+Claude-Sonnet-4.5 & 5.07M$\pm$0.12M & 81.73K$\pm$3.45K \\
& Codex+GPT-5.2 & 8.63M$\pm$1.81M & 112.28K$\pm$17.22K \\
\midrule
\multirow{4}{*}{\shortstack{Bioimage\\Visualization}}
& BioImage-Agent+GPT-5.2 & {\bf 931.66K$\pm$186.31K} & {\bf 6.26K$\pm$1.47K} \\
& BioImage-Agent+Claude-Sonnet-4.5  & 1.58M$\pm$0.02M & 18.47K$\pm$1.28K \\
& Claude-Code+Claude-Sonnet-4.5 & 8.60M$\pm$0.17M & 125.66K$\pm$2.43K \\
& Codex+GPT-5.2 & 10.36M$\pm$3.76M & 104.83K$\pm$25.78K \\
\midrule
\multirow{4}{*}{\shortstack{Topology\\Visualization}}
& TopoPilot+GPT-5.2    & {\bf 394.00K$\pm$6.06K} & {\bf 2.89K$\pm$0.17K} \\
& TopoPilot+Claude-Sonnet-4.5 & 3.74M$\pm$3.67M & 57.93K$\pm$32.35K \\
& Claude-Code+Claude-Sonnet-4.5 & 17.26M$\pm$1.87M & 172.37K$\pm$18.51K \\
& Codex+GPT-5.2 & 46.04M$\pm$4.48M & 193.58K$\pm$27.62K \\
\midrule
\multirow{4}{*}{\shortstack{Object\\Identification}}
& ParaView-MCP+GPT-5.2     & {\bf 4.12M$\pm$0.39M} & {\bf 24.58K$\pm$1.20K} \\
& ParaView-MCP+Claude-Sonnet-4.5  & 11.42M$\pm$0.49M & 119.24K$\pm$4.65K \\
& Claude-Code+Claude-Sonnet-4.5 & 16.83M$\pm$0.28M & 273.90K$\pm$22.71K \\
& Codex+GPT-5.2 & 38.43M$\pm$3.55M & 344.27K$\pm$14.95K \\
\bottomrule
\end{tabular}
\end{adjustbox}
	\vspace{-0.1in}
\end{table}

\vspace{-0.075in}
\section{Baselines and Benchmark Results}

\subsection{Baseline Agents and Setup}

The current benchmark includes task suites covering core ParaView-based SciVis operations, such as volume rendering, isosurface extraction, flow visualization, scientific plotting, and data-driven analysis. It also includes domain-specific suites for molecular visualization with GMX and VMD, bioimage visualization with napari, and topology visualization with TTK. Finally, we include a suite of object identification tasks that load anonymized volume datasets, adjust transfer functions, and identify unknown objects. This task suite is designed to evaluate whether agents can both execute the visualization workflow and use visual evidence to infer which object is being shown, thereby probing their vision-based reasoning capabilities.

For ParaView visualization tasks, we evaluated ChatVis~\cite{peterka2025chatvis} and ParaView-MCP~\cite{liu2025paraview}. 
For molecular visualization tasks, we evaluated GMX-VMD-MCP~\cite{egtai2025gmxvmdmcp}. 
For biomimage visualization tasks, we evaluated BioImage-Agent~\cite{llnl2025bioimageagent}. 
For topology visualization tasks, we evaluated TopoPilot~\cite{gorski2026topopilot}.
For object identification tasks, we evaluate ParaView-MCP, which employs a perception–action loop to iteratively refine visualizations and progressively build an understanding of the data.
All agents were tested with two backbone LLMs, GPT-5.2 and Claude-Claude-Sonnet-4.5.
Because each domain requires specialized software environments and existing SciVis agents typically support only a single toolchain, no individual agent can complete tasks across all suites. Each agent was therefore evaluated only on the task suite compatible with its supported environment.

To provide a cross-domain reference point, we additionally evaluated two general-purpose coding agents, Claude Code~\cite{anthropic2025claudecode} and Codex~\cite{openai2025codex}, under a minimal-tool setting. Specifically, both agents had access only to the underlying visualization engines (ParaView, napari, VMD, and TTK, pre-installed in a conda environment). They were required to complete benchmark tasks by generating executable scripts that invoked these tools. No additional API references, agent skills, or online access to documents were provided; they were run in the built-in sandbox mode with full automation.

\vspace{-0.05in}
\subsection{Benchmark Results and Analysis}

According to our human–LLM alignment study, Claude-Opus-4.6 is the most appropriate judge, both in alignment with human SciVis experts and in evaluation stability. More importantly, Opus-4.6 is not used as the task model, so we can entirely separate the model from task execution during evaluation. We therefore reported the performance of baseline SciVis agents and the general-purpose coding agent across all five task suites of SciVisAgentBench. Each experimental setting was repeated three times, and we reported the mean and standard deviation of the overall score and task completion rate.

The results evaluated using Claude-Opus-4.6 are reported in Table~\ref{tab:main_results_all_parts_opus46}. For completeness, we also report the same results evaluated by GPT-5.2 (Table~\ref{tab:main_results_all_parts_gpt52} in the appendix). In these tables, the completion rate indicates the proportion of runs that finish without explicit execution errors. In contrast, pass metrics require the agent to produce a valid output (e.g., saving results to the expected location and generating a non-empty visualization). Topology visualization results are reported only once in Table~\ref{tab:main_results_all_parts_opus46}, since this task suite uses rule-based evaluation and does not rely on an LLM judge.
Both tables show similar performance trends: general-purpose coding agents outperform most specialized SciVis agents across most task suites. However, the evaluation using Claude-Opus-4.6 yields slightly lower mean scores and smaller variances, suggesting that it provides stricter, more stable judgments than GPT-5.2. This observation is consistent with the findings reported in Section~\ref{subsec:outcome_valid}.

\hot{
Ground-truth and agent-generated outcomes for representative examples, along with their task descriptions, goal-level rubrics, and per-goal evaluation scores, are provided in Appendix~\ref{appendix:qualitative-examples}. 
These examples show how numerical differences correspond to concrete failures such as missing streamlines, incorrect chart construction, poor isosurface extraction, or mismatched color mappings. They highlight that SciVisAgentBench does not rely solely on visual similarity; its evaluation is grounded in expert-authored rubrics and task-specific criteria that assess whether the generated visualization meets the intended scientific and analytical objectives. However, vision-based LLM judges remain imperfect, particularly for subjective assessments and higher-level scientific interpretation, motivating our use of complementary deterministic evaluators and the human-LLM alignment study described in Section~\ref{subsec:outcome_valid}.}

In Table~\ref{tab:paraview_image_metrics}, we further report image-based evaluation metrics for ParaView visualization tasks. These metrics are scaled variants of the aggregated PSNR, SSIM, and LPIPS scores (see Section~\ref{subsec:deterministic-evaluator}). Because these metrics require ground-truth visualizations rendered under controlled conditions, they were computed only for the ParaView visualization task suite.
The results are consistent with the LLM-based evaluation. General-purpose coding agents achieve better image similarity than the specialized baselines. In particular, Claude Code and Codex obtain much higher $\mathrm{PSNR}_{\text{scaled}}$ and $\mathrm{SSIM}_{\text{scaled}}$, together with much lower $\mathrm{LPIPS}_{\text{scaled}}$, and perform comparably across all three image metrics, indicating that their outputs are both structurally and perceptually closer to the ground-truth visualizations.

In Table~\ref{tab:token_cost_all_parts}, we report the input and output token usage across all experiments. To ensure consistent accounting across experiments, token counts are reported without input caching, treating cached tokens as regular input tokens. The results show that general-purpose coding agents typically consume more tokens than specialized SciVis agents, reflecting the additional interactions needed for environment exploration, code generation, and iterative debugging. In contrast, MCP-based agents often achieve lower token usage because their predefined tool pipelines reduce the need for extensive exploration.

\vspace{-0.05in}
\subsection{Agent Choices and Tradeoffs}

Overall, the general-purpose coding agents outperform most specialized SciVis agents by a large margin across several task suites. In addition, when comparing the same SciVis agent across different backbone models, Claude-Sonnet-4.5 generally performs better than GPT-5.2.
However, improved agentic performance often comes at the cost of higher token usage and longer execution time, as shown in Table~\ref{tab:token_cost_all_parts}. In visualization and data analysis scenarios with limited context, LLM agents must interact dynamically with the environment and progressively refine their solutions through trial and error. Despite these challenges, general-purpose coding agents can generate useful executable code under such conditions, though often at the cost of higher token usage.

Nevertheless, general-purpose coding agents also exhibit several failure modes. They may spend many turns probing the environment to determine which libraries or tools are available, even when the environment is already properly configured. They may also misuse visualization APIs or capture incorrect outputs (e.g., Codex takes screenshots of the entire napari GUI rather than the visualization viewport), increasing interaction overhead and potentially leading to low-quality results.

In contrast, MCP-based agents emphasize efficiency and reliability through predefined tool pipelines. They remain valuable in scenarios where tasks are well-defined and predictable. In such settings, MCP pipelines provide reliable and reusable solutions with minimal token usage. For example, in the bioimage visualization task suite of our benchmark, BioImage-Agent outperforms both Claude Code and Codex while using significantly fewer tokens, thanks to its MCP tools, which provide good coverage of the required task operations. In contrast, general-purpose agents, especially Codex, often struggle with napari because they lack task-specific knowledge of the specific software version. However, when tasks exceed their predefined capabilities, MCP agents may fail or revert to more exploratory behaviors, akin to those of general-purpose coding agents.

\vspace{-0.05in}
\section{Call for Participation}


SciVisAgentBench is intended as a starting point rather than a complete coverage of the SciVis field. To foster continuous growth, we invite participation in two key areas, as outlined below.
First, the benchmark should expand in breadth. Although the current release covers a diverse set of application domains and data types, many important areas of SciVis remain underrepresented. Expanding the benchmark with new datasets, domains, and visualization scenarios will improve coverage and better reflect real-world practice.
Second, the benchmark should go deeper into complexity levels. Beyond isolated tasks, future development should emphasize long-horizon workflows that more closely resemble real scientific processes, including environment setup, iterative analysis, visualization design, and result reporting. For example, whether LLM agents can complete end-to-end workflows comparable to an IEEE VIS contest designed for human participants.

We invite the VIS community to contribute to both directions. Researchers and practitioners can submit new datasets, cases, and evaluation scenarios through the infrastructure we have built, following the provided templates and taxonomy to ensure consistency and reproducibility.
%
To support this effort, we outline several concrete actions. We plan to organize workshops focused on SciVis agent evaluation, provide tutorials for developing and contributing benchmark cases, and host benchmark-driven challenges. In particular, we plan to organize a contest where agents, rather than humans, participate under a shared evaluation protocol.

\vspace{-0.075in}
\section{Conclusions and Future Work}

As we witness the shift from generative AI~\cite{Wang-TVCG23} to agentic AI in SciVis, SciVisAgentBench addresses a critical gap in agentic visualization by providing a principled, reproducible framework for evaluating LLM-based agentic systems on complex, real-world scientific data analysis and visualization tasks. The 108 cases span multiple dimensions, organized along a taxonomy of application domains, data types, complexity levels, and visualization operations. Through a comprehensive validity study with 12 SciVis experts, we validated that state-of-the-art MLLMs can reliably approximate human expert judgment (Pearson correlation of 0.806 with Claude-Opus-4.6), unlocking scalable, automated assessment for traditionally subjective visualization tasks. 

Our baseline evaluations expose a compelling dichotomy in the current ecosystem: specialized, tool-integrated agents excel in efficiency and reliability within predefined pipelines, while general-purpose coding agents demonstrate superior adaptability and overall success across diverse domains. Most significantly, our analysis reveals a promising direction that may resolve this dichotomy: general-purpose coding agents augmented with domain-specific skills combine the best of both paradigms \hot{(refer to Table~\ref{tab:main_results_bioimage_gpt52} in the appendix and Appendix~\ref{appendix:agent_skill})}. 

SciVisAgentBench itself serves dual purposes: as a standardized \textit{yardstick} for measuring progress and as a diagnostic tool for identifying capability gaps and guiding architectural choices. 
\hot{Our results suggest that outcome-centric evaluation provides a practical foundation for assessing SciVis tasks with explicit outcomes, while the validated LLM judging methodology offers a scalable path toward automated evaluation. Meanwhile, the current benchmark intentionally focuses on structured workflows with verifiable outcomes. It does not yet address open-ended exploration, long-horizon scientific discovery, or tasks with multiple equally valid visualization designs.} Designed as an extensible, living benchmark with established contribution infrastructure, SciVisAgentBench moves beyond static evaluation toward continuous evolution that tracks both agent capabilities and emerging SciVis needs (through potentially versioned releases).

Looking ahead, several directions emerge. In the near term, we welcome community contributions to expand coverage across underrepresented application domains, data types, and visualization operations through our established submission pipeline. The benchmark can serve as an environment that enables agents to iterate on and improve their performance on general data analysis tasks. In the medium term, we envision extending from task-level assessment to more long-horizon scientific workflows that reveal data insights. Ultimately, we hope SciVisAgentBench will serve as a foundation for principled, measurable progress in autonomous scientific data analysis and visualization, enabling systematic comparison, surfacing open challenges, and accelerating the development of the next generation of reliable, capable agents that can meaningfully assist domain scientists in their day-to-day analytical workflows.

\vspace{-0.1in}
\acknowledgments{
This work was performed under the auspices of the U.S.\ Department of Energy by Lawrence Livermore National Laboratory under contract DE-AC52-07NA27344, reviewed and released under LLNL-CONF-2017490. The work is partially funded by DOE ECRP 51917/SCW1885 and ASCR DE-SC0023145, and the U.S.\ National Science Foundation IIS-2101696, OAC-2104158, IIS-2401144, and IIS-2550610. 
The authors thank the anonymous reviewers for their insightful comments.}

\vspace{-0.1in}
\bibliographystyle{abbrv-doi-hyperref-narrow}
\bibliography{refs-abbv}

\appendix 
\crefalias{section}{appendix} 

\newpage
\clearpage

\setcounter{section}{0}
\setcounter{figure}{0}
\setcounter{table}{0}


\section{Benchmark Taxonomy}
\label{appendix:taxonomy}

We list the taxonomy used in SciVisAgentBench, including application domain, data type, complexity level, and visualization operation.

\vspace{-0.05in}
\subsection{Application Domain Taxonomy}

The benchmark covers the following eight application domains: Astronomy, Biology, Chemistry, Earth System Science, Mathematics, Medical Science, Physics, and Others.

\vspace{-0.05in}
\subsection{Data Type Taxonomy}

The benchmark spans the following five data types:
Scalar field, Vector field, Tensor field, Multivariate, and Time-varying.

\vspace{-0.05in}
\subsection{Complexity Level Taxonomy}

Each benchmark case is categorized into one of the three complexity levels based on the highest-level procedural scope:
\begin{myitemize}
\vspace{-0.05in}
\item \textbf{Operation}: a single, focused visualization procedure.
\item \textbf{Task}: multiple coordinated operations toward a specific objective.
\item \textbf{Workflow}: a multi-stage visualization pipeline or complex analytical process.
\vspace{-0.05in}
\end{myitemize}
In practice, we report only task- and workflow-level entries as benchmark cases;  operations serve as the building blocks within these higher-level categories.

\vspace{-0.05in}
\subsection{Visualization Operation Taxonomy}
\label{subsec:app-vot}

A visualization operation describes the primary analytical or rendering actions performed. Each benchmark case is tagged with one or more of the following 15 operations.

\vspace{-0.05in}
\subsubsection{Color \& Opacity Mapping}

This operation assigns visual attributes based on data values. It includes colormap assignment, opacity mapping, texture application, material properties, and contrast adjustments. In molecular visualization, coloring by element, residue, or other attributes falls into this category. The operation defines visual encoding rules rather than geometry.

\vspace{-0.05in}
\subsubsection{Data Sampling \& Resolution Control}

This operation modifies data density or sampling resolution while preserving overall semantics. It includes subsampling, decimation, resampling onto different grids, probing fields at user-specified locations, and generating reduced representations for performance or clarity. Unlike subsetting, this operation controls representation fidelity rather than selecting regions of interest.

\vspace{-0.05in}
\subsubsection{Data Smoothing \& Filtering}

This operation reduces noise or enhances data quality without changing geometry or topology. It includes Laplacian or Gaussian smoothing, noise reduction, interpolation-based gap filling, edge enhancement, feature sharpening, outlier removal, and statistical filtering.

\vspace{-0.05in}
\subsubsection{Data Subsetting \& Extraction}

This operation isolates spatial regions or value-based subsets from a dataset. It includes clipping by plane, box, or sphere; extracting a volume of interest; selecting by point or cell IDs; thresholding by scalar range; isocontouring or isosurfacing to extract constant-value boundaries; and connectivity filtering to isolate connected components. The defining characteristic is reducing the dataset to a subset based on spatial location or attribute value.

\vspace{-0.05in}
\subsubsection{Dataset Restructuring}

This operation reorganizes datasets at the structural or container level. It includes merging datasets, appending blocks, partitioning datasets, and converting between data formats. The defining characteristic is the modification of the dataset organization rather than geometry or attributes.

\vspace{-0.05in}
\subsubsection{Feature Identification \& Segmentation}

This operation detects, extracts, or labels meaningful structures within data. It includes connected component labeling, object detection, region segmentation, topology-based feature classification, and structure identification. The output is a set of labeled objects or regions.

\vspace{-0.05in}
\subsubsection{Field Computation}

This operation derives new scalar, vector, or tensor fields from existing attributes. It includes gradient, divergence, curl, vorticity, curvature, arithmetic field operations, vector magnitude computation, statistical summaries, tensor eigendecomposition, interpolation, and distance calculations. Particle tracing and critical point detection are also included.

\vspace{-0.05in}
\subsubsection{Geometric \& Topological Transformation}

This operation modifies spatial structure or connectivity without deriving new field values. It includes translation, rotation, scaling, deformation, warping, triangulation, mesh refinement or coarsening, and boundary extraction. The focus is on the structural transformation of existing data.

\vspace{-0.05in}
\subsubsection{Glyph \& Marker Placement}

This operation places discrete visual symbols at data locations to encode local attributes. It includes oriented vector glyphs, scaled glyphs, tensor glyphs, and seed markers. Molecular representations such as spheres, sticks, cartoons, and surfaces are also included. The defining feature is symbol-based encoding of localized data values.

\vspace{-0.05in}
\subsubsection{Plot \& Chart Generation}

This operation produces 2D statistical visualizations from data attributes. It includes histograms, line plots, bar charts, and other standard statistical graphics. The output is a 2D chart rather than a spatial visualization.

\vspace{-0.05in}
\subsubsection{Scientific Insight Derivation}

This operation uses visualization or analysis results to answer domain-specific scientific questions. Examples include assessing molecular penetration, verifying anatomical structures, or validating simulation plausibility. The defining characteristic is that the output is knowledge or a decision rather than a visualization artifact.

\vspace{-0.05in}
\subsubsection{Surface \& Contour Extraction}

This operation generates geometric primitives representing boundaries or cross-sections within volumetric data. It includes isosurface generation, contour extraction, and planar slicing. 
The key distinction is that new geometry is produced to represent feature boundaries or cross-sections.

\vspace{-0.05in}
\subsubsection{Temporal Processing}

This operation explicitly uses the time dimension in time-varying scalar fields or unsteady vector fields. It includes temporal interpolation, temporal aggregation, pathline/streakline/timeline computation, flow map computation, and timestep navigation.

\vspace{-0.05in}
\subsubsection{View \& Camera Control}

This operation manipulates the viewing context rather than the data. It includes camera rotation, zooming, panning, viewpoint presets, lighting adjustment, switching between 2D and 3D modes, and camera reset. Molecular view centering and orientation adjustments are also included.

\vspace{-0.05in}
\subsubsection{Volume Rendering}

This operation visualizes volumetric data directly without reducing it to surfaces. It includes ray casting, splatting, and transfer function design for mapping scalar values to color and opacity. Multi-channel rendering and X-ray style transparency are also included. The defining property is that the full volume contributes to the final image.

\vspace{-0.05in}
\section{Benchmark Cases}
\label{appendix:cases}

Table~\ref{tab:all_cases} provides the complete list of 108 benchmark cases in SciVisAgentBench.

\vspace{-0.05in}
\section{Additional Results}
\label{appendix:additional}

Table~\ref{tab:main_results_all_parts_gpt52} reports the overall benchmark performance using GPT-5.2 as the LLM judge.
Table~\ref{tab:main_results_bioimage_gpt52} further reports the full benchmark performance of Claude Code on the bioimage visualization task suite with different backbones and skill settings.

\clearpage
\onecolumn
{\small
\begin{longtable}{p{0.18\textwidth} p{0.12\textwidth} p{0.08\textwidth} p{0.06\textwidth} p{0.44\textwidth}}
\caption{The complete list of all 108 benchmark cases in SciVisAgentBench, including their application domains, data types, complexity levels, and visualization operations. Cases are ordered alphabetically based on case names.}
\vspace{-0.1in}
\label{tab:all_cases} \\
\toprule
\textbf{Case Name} & \textbf{Domain} & \textbf{Type} & \textbf{Level} & \textbf{Operations} \\
\midrule
\endfirsthead

\toprule
\textbf{Case Name} & \textbf{Domain} & \textbf{Type} & \textbf{Level} & \textbf{Operations} \\
\midrule
\endhead

\bottomrule
\endfoot

1CRN\_color-by-charge & Biology & Scalar Field & Task & Color \& Opacity Mapping; Glyph \& Marker Placement; View \& Camera Control \\
1CRN\_contacts & Biology & Scalar Field & Task & Field Computation \\
1CRN\_cpk-coloring & Biology; Chemistry & Scalar Field & Task & Color \& Opacity Mapping; Glyph \& Marker Placement; View \& Camera Control \\
1CRN\_distance-angles & Biology & Scalar Field & Task & Field Computation \\
1CRN\_licorice & Biology & Scalar Field & Task & Glyph \& Marker Placement; View \& Camera Control \\
1CRN\_radius-gyration & Biology & Scalar Field & Task & Field Computation \\
1CRN\_rmsd-rmsf & Biology & Scalar Field & Task & Field Computation \\
1CRN\_select-aromatic & Biology & Scalar Field & Task & Color \& Opacity Mapping; Glyph \& Marker Placement; View \& Camera Control \\
1CRN\_select-carbon & Biology & Scalar Field & Task & Color \& Opacity Mapping; Glyph \& Marker Placement; View \& Camera Control \\
1CRN\_select-oxygen & Biology & Scalar Field & Task & Color \& Opacity Mapping; Glyph \& Marker Placement; View \& Camera Control \\
ABC & Physics & Vector Field & Workflow & Field Computation; Glyph \& Marker Placement; Color \& Opacity Mapping; View \& Camera Control \\
aneurysm & Medical Science & Scalar Field & Task & Volume Rendering; Color \& Opacity Mapping \\
argon-bubble & Chemistry & Scalar Field & Task & Volume Rendering; Color \& Opacity Mapping; View \& Camera Control \\
backpack & Others & Scalar Field & Task & Volume Rendering; Color \& Opacity Mapping \\
bcell-receptor\_camera-ops & Biology & Multivariate & Task & Color \& Opacity Mapping; View \& Camera Control \\
bcell-receptor\_cell-counting & Biology & Scalar Field & Task & Color \& Opacity Mapping; View \& Camera Control; Scientific Insight Derivation \\
bcell-receptor\_cell-trace & Biology & Scalar Field & Task & Glyph \& Marker Placement; View \& Camera Control \\
bcell-receptor\_ingest-labels & Biology & Multivariate & Task & Feature Identification \& Segmentation; Scientific Insight Derivation \\
bcell-receptor\_ingest-points & Biology & Multivariate & Task & Glyph \& Marker Placement; Scientific Insight Derivation \\
bcell-receptor\_ingest-shapes & Biology & Multivariate & Task & Glyph \& Marker Placement; Scientific Insight Derivation \\
bcell-receptor\_multi-channel & Biology & Time-varying; Multivariate & Workflow & Color \& Opacity Mapping; Temporal Processing; View \& Camera Control; Scientific Insight Derivation \\
Bernard & Physics & Vector Field & Workflow & Field Computation; Glyph \& Marker Placement; Color \& Opacity Mapping; View \& Camera Control \\
blunt-fin & Physics & Scalar Field & Task & Volume Rendering; Color \& Opacity Mapping \\
bonsai\_dvr & Biology & Scalar Field & Task & Volume Rendering; Color \& Opacity Mapping \\
bonsai\_interaction & Biology & Scalar Field & Task & Volume Rendering; Color \& Opacity Mapping; View \& Camera Control \\
Boston-teapot & Others & Scalar Field & Task & Volume Rendering; Color \& Opacity Mapping \\
brain & Medical Science; Mathematics & Tensor Field & Workflow & Field Computation; Feature Identification \& Segmentation; Scientific Insight Derivation \\
bunny & Others & Scalar Field & Task & Volume Rendering; Color \& Opacity Mapping \\
can\_color-blocks & Others & Scalar Field & Task & Color \& Opacity Mapping; View \& Camera Control \\
can\_sub-time-series & Others & Time-varying & Task & Temporal Processing; Volume Rendering; View \& Camera Control \\
can\_temporal-animation & Others & Time-varying & Task & Temporal Processing; Color \& Opacity Mapping; View \& Camera Control \\
carp\_dvr & Biology & Scalar Field & Task & Volume Rendering; Color \& Opacity Mapping \\
carp\_analysis & Biology & Scalar Field & Workflow & Volume Rendering; Color \& Opacity Mapping; View \& Camera Control; Scientific Insight Derivation \\
chameleon\_isosurface & Biology & Scalar Field & Task & Surface \& Contour Extraction; Color \& Opacity Mapping; View \& Camera Control \\
climate & Earth System Science & Vector Field & Workflow & Field Computation; Glyph \& Marker Placement; Color \& Opacity Mapping; View \& Camera Control \\
cortical-neurons\_annotation & Biology & Multivariate & Task & Glyph \& Marker Placement \\
cortical-neurons\_figure-recreation & Biology & Multivariate & Task & Color \& Opacity Mapping; View \& Camera Control \\
cortical-neurons\_statistics & Biology & Multivariate & Task & Field Computation; Scientific Insight Derivation \\
crayfish\_streamline & Physics & Vector Field & Workflow & Field Computation; Glyph \& Marker Placement; Color \& Opacity Mapping; View \& Camera Control \\
curved-membrane & Biology & Scalar Field & Workflow & Glyph \& Marker Placement; Color \& Opacity Mapping; View \& Camera Control; Scientific Insight Derivation \\
cylinder & Physics; Mathematics & Scalar Field & Workflow & Field Computation; Feature Identification \& Segmentation; Data Smoothing \& Filtering; Scientific Insight Derivation \\
disk-out-ref\_line-plot & Others & Scalar Field & Task & Plot \& Chart Generation; Color \& Opacity Mapping; View \& Camera Control \\
dragon & Mathematics & Scalar Field & Workflow & Field Computation; Feature Identification \& Segmentation \\
engine\_dvr & Others & Scalar Field & Task & Volume Rendering; Color \& Opacity Mapping \\
engine\_interaction & Others & Scalar Field & Task & Volume Rendering; Color \& Opacity Mapping; View \& Camera Control \\
foot\_dvr & Medical Science & Scalar Field & Task & Volume Rendering; Color \& Opacity Mapping \\
foot\_analysis & Medical Science & Scalar Field & Workflow & Volume Rendering; Color \& Opacity Mapping; View \& Camera Control; Scientific Insight Derivation \\
frog & Biology & Scalar Field & Task & Volume Rendering; Color \& Opacity Mapping \\
fuel & Physics & Scalar Field & Task & Volume Rendering; Color \& Opacity Mapping \\
hydrogen-atom & Physics & Scalar Field & Task & Volume Rendering; Color \& Opacity Mapping \\
Isabel & Earth System Science; Mathematics & Scalar Field & Workflow & Field Computation; Feature Identification \& Segmentation; Data Smoothing \& Filtering; Scientific Insight Derivation \\
lobster\_dvr & Biology & Scalar Field & Task & Volume Rendering; Color \& Opacity Mapping \\
lobster\_isosurface & Biology & Scalar Field & Workflow & Surface \& Contour Extraction; Color \& Opacity Mapping; View \& Camera Control; Scientific Insight Derivation \\
materials & Others & Scalar Field & Task & Data Subsetting \& Extraction; View \& Camera Control \\
mhd-magfield\_streamribbon & Physics & Vector Field & Workflow & Field Computation; Glyph \& Marker Placement; Color \& Opacity Mapping; View \& Camera Control \\
mhd-turbulence\_pathline & Physics & Time-varying; Vector Field & Workflow & Temporal Processing; Glyph \& Marker Placement; Color \& Opacity Mapping; View \& Camera Control \\
mhd-turbulence\_pathribbon & Physics & Time-varying; Vector Field & Workflow & Temporal Processing; Glyph \& Marker Placement; Data Smoothing \& Filtering; Geometric \& Topological Transformation; Color \& Opacity Mapping; View \& Camera Control \\
mhd-turbulence\_streamline & Physics & Vector Field & Workflow & Field Computation; Glyph \& Marker Placement; Color \& Opacity Mapping; View \& Camera Control \\
miranda & Physics & Scalar Field & Task & Volume Rendering; Color \& Opacity Mapping; View \& Camera Control \\
ml\_dvr & Others & Scalar Field & Task & Volume Rendering; Color \& Opacity Mapping; View \& Camera Control \\
ml\_isosurface & Others & Scalar Field & Task & Surface \& Contour Extraction; Color \& Opacity Mapping; View \& Camera Control \\
ml\_slice-contour & Others & Scalar Field & Task & Data Subsetting \& Extraction; Surface \& Contour Extraction; Color \& Opacity Mapping; View \& Camera Control \\
molecule & Chemistry; Mathematics & Scalar Field & Workflow & Field Computation; Feature Identification \& Segmentation \\
moons & Mathematics & Scalar Field & Workflow & Field Computation; Feature Identification \& Segmentation; Data Smoothing \& Filtering \\
mri-ventricles & Medical Science & Scalar Field & Task & Volume Rendering; Color \& Opacity Mapping \\
mri-woman & Medical Science & Scalar Field & Task & Volume Rendering; Color \& Opacity Mapping \\
mrt-angio & Medical Science & Scalar Field & Task & Volume Rendering; Color \& Opacity Mapping \\
neocortical-layer-1-axons & Biology & Scalar Field & Task & Volume Rendering; Color \& Opacity Mapping \\
noisy-terrain & Mathematics & Scalar Field & Workflow & Field Computation; Feature Identification \& Segmentation; Data Smoothing \& Filtering \\
nucleon & Physics & Scalar Field & Task & Volume Rendering; Color \& Opacity Mapping \\
ocean & Earth System Science; Mathematics & Tensor Field & Workflow & Field Computation; Feature Identification \& Segmentation; Scientific Insight Derivation \\
pancreas & Medical Science & Scalar Field & Task & Volume Rendering; Color \& Opacity Mapping \\
points\_surf-clip & Others & Scalar Field & Task & Data Subsetting \& Extraction; Glyph \& Marker Placement; View \& Camera Control \\
QMCPACK & Physics; Mathematics & Scalar Field & Workflow & Field Computation; Feature Identification \& Segmentation; Scientific Insight Derivation \\
ras-raf-membrane & Biology & Scalar Field & Workflow & Glyph \& Marker Placement; Color \& Opacity Mapping; View \& Camera Control; Scientific Insight Derivation \\
Richtmyer & Physics & Scalar Field & Task & Volume Rendering; Color \& Opacity Mapping; View \& Camera Control \\
rings\_import-gltf & Others & Scalar Field & Task & Dataset Restructuring; View \& Camera Control \\
rotstrat & Physics & Scalar Field & Task & Volume Rendering; Color \& Opacity Mapping; View \& Camera Control \\
rti-velocity\_glyph & Physics & Vector Field & Workflow & Data Subsetting \& Extraction; Glyph \& Marker Placement; Color \& Opacity Mapping; View \& Camera Control \\
rti-velocity\_slices & Physics & Vector Field & Workflow & Data Subsetting \& Extraction; Color \& Opacity Mapping; View \& Camera Control \\
rti-velocity\_streakline & Physics & Time-varying; Vector Field & Workflow & Temporal Processing; Glyph \& Marker Placement; Color \& Opacity Mapping; View \& Camera Control \\
shrink-sphere & Others & Scalar Field & Task & Geometric \& Topological Transformation; View \& Camera Control \\
silicium & Physics & Scalar Field & Task & Volume Rendering; Color \& Opacity Mapping \\
skull & Medical Science & Scalar Field & Task & Volume Rendering; Color \& Opacity Mapping \\
solar-plume & Astronomy & Scalar Field & Workflow & Volume Rendering; Color \& Opacity Mapping; View \& Camera Control \\
statue-leg & Others & Scalar Field & Task & Volume Rendering; Color \& Opacity Mapping \\
stent & Medical Science & Scalar Field & Task & Volume Rendering; Color \& Opacity Mapping \\
stream-glyph & Others & Vector Field & Task & Field Computation; Glyph \& Marker Placement; Color \& Opacity Mapping; View \& Camera Control \\
supernova & Astronomy & Scalar Field & Task & Volume Rendering; Color \& Opacity Mapping \\
supernova\_isosurface & Astronomy & Scalar Field & Workflow & Surface \& Contour Extraction; Data Subsetting \& Extraction; Color \& Opacity Mapping; View \& Camera Control \\
supernova\_streamline & Astronomy & Vector Field & Workflow & Field Computation; Glyph \& Marker Placement; Color \& Opacity Mapping; View \& Camera Control \\
tgc-velocity\_contour & Physics & Vector Field & Workflow & Surface \& Contour Extraction; Field Computation; Color \& Opacity Mapping; View \& Camera Control \\
Tangaroa\_streamribbon & Physics & Vector Field & Workflow & Field Computation; Glyph \& Marker Placement; Color \& Opacity Mapping; View \& Camera Control \\
tooth & Medical Science & Scalar Field & Task & Volume Rendering; Color \& Opacity Mapping \\
tooth\_isosurface & Biology & Scalar Field & Task & Surface \& Contour Extraction; View \& Camera Control \\
tornado\_dvr & Earth System Science & Scalar Field & Task & Volume Rendering; Color \& Opacity Mapping \\
tornado\_glyph-streamline & Earth System Science & Vector Field & Workflow & Field Computation; Glyph \& Marker Placement; Color \& Opacity Mapping; View \& Camera Control \\
trajectory-inspection & Biology & Time-varying & Workflow & Temporal Processing; Plot \& Chart Generation; View \& Camera Control; Scientific Insight Derivation \\
transparent-bg & Others & Scalar Field & Task & View \& Camera Control \\
two-swirls\_streamribbon & Physics & Vector Field & Workflow & Field Computation; Glyph \& Marker Placement; Color \& Opacity Mapping; View \& Camera Control \\
vis-male & Medical Science & Scalar Field & Task & Volume Rendering; Color \& Opacity Mapping \\
vortex & Physics & Scalar Field & Task & Volume Rendering; Color \& Opacity Mapping; View \& Camera Control \\
wavelet\_camera & Others & Scalar Field & Task & View \& Camera Control \\
wavelet\_chart & Others & Scalar Field & Task & Plot \& Chart Generation; Volume Rendering; Color \& Opacity Mapping; View \& Camera Control \\
wavelet\_colormap & Others & Scalar Field & Task & Color \& Opacity Mapping; View \& Camera Control \\
wavelet\_export-gltf & Others & Scalar Field & Task & Surface \& Contour Extraction; Dataset Restructuring; View \& Camera Control \\
wavelet\_export-ply & Others & Scalar Field & Task & Surface \& Contour Extraction; Dataset Restructuring \\
wavelet\_histogram & Others & Scalar Field & Task & Plot \& Chart Generation; Color \& Opacity Mapping; View \& Camera Control \\
\end{longtable}
}

\begin{table*}[t]
\centering
\small
\caption{Benchmark performance across four task suites of SciVisAgentBench using {\bf GPT-5.2} as the LLM judge for evaluation. 
Scores and completion rates are reported as mean$\pm$std across three repeated trials. 
Topology visualization results are excluded because they use rule-based evaluation and do not require an LLM judge.}
\label{tab:main_results_all_parts_gpt52}
\vspace{-0.1in}
\begin{adjustbox}{width=\textwidth}
\begin{tabular}{clcccccccc}
\toprule
Task Suite & Setting & Overall Score $\uparrow$ & Completion Rate $\uparrow$ & pass@1 $\uparrow$ & pass@2 $\uparrow$ & pass@3 $\uparrow$ & pass$\textasciicircum1$ $\uparrow$ & pass$\textasciicircum2$ $\uparrow$ & pass$\textasciicircum3$ $\uparrow$ \\
\midrule
\multirow{6}{*}{\shortstack{ParaView\\Visualization}}
& ChatVis+GPT-5.2 & 32.77$\pm$0.25 & 47.92$\pm$2.08 & 0.389 & 0.569 & 0.688 & 0.389 & 0.208 & 0.146 \\
& ChatVis+Claude-Sonnet-4.5 & 39.08$\pm$3.50 & 56.25$\pm$4.66 & 0.479 & 0.562 & 0.604 & 0.479 & 0.396 & 0.354 \\
& ParaView-MCP+GPT-5.2 & 27.47$\pm$4.12 & 51.39$\pm$6.70 & 0.299 & 0.368 & 0.417 & 0.299 & 0.229 & 0.208 \\
& ParaView-MCP+Claude-Sonnet-4.5 & 29.80$\pm$9.46 & 59.72$\pm$15.07 & 0.285 & 0.389 & 0.479 & 0.285 & 0.181 & 0.167 \\
& Claude-Code+Claude-Sonnet-4.5 & \textbf{64.87$\pm$0.83} & \textbf{98.61$\pm$2.41} & \textbf{0.778} & 0.861 & 0.896 & \textbf{0.778} & \textbf{0.694} & \textbf{0.646} \\
& Codex+GPT-5.2 & 63.13$\pm$1.34 & 97.22$\pm$2.41 & \textbf{0.778} & \textbf{0.910} & \textbf{0.985} & \textbf{0.778} & 0.646 & 0.562 \\
\midrule
\multirow{4}{*}{\shortstack{Molecular\\Visualization}}
& GMX-VMD-MCP+GPT-5.2 & 34.10$\pm$15.59 & 56.41$\pm$8.88 & 0.513 & 0.667 & 0.769 & 0.513 & 0.359 & 0.308 \\
& GMX-VMD-MCP+Claude-Sonnet-4.5 & 58.27$\pm$5.46 & 92.31$\pm$7.69 & 0.821 & 0.872 & 0.923 & 0.821 & 0.769 & 0.769 \\
& Claude-Code+Claude-Sonnet-4.5 & 64.47$\pm$8.41 & 94.87$\pm$4.44 & \textbf{0.923} & \textbf{0.974} & \textbf{1.000} & \textbf{0.923} & \textbf{0.872} & \textbf{0.846} \\
& Codex+GPT-5.2 & \textbf{69.27$\pm$4.21} & \textbf{97.44$\pm$4.44} & 0.897 & \textbf{0.974} & \textbf{1.000} & 0.897 & 0.821 & 0.769 \\
\midrule
\multirow{4}{*}{\shortstack{Bioimage\\Visualization}}
& BioImage-Agent+GPT-5.2 & \textbf{61.47$\pm$14.89} & 81.82$\pm$0.00 & \textbf{0.667} & \textbf{0.788} & \textbf{0.818} & \textbf{0.667} & 0.545 & 0.455 \\
& BioImage-Agent+Claude-Sonnet-4.5 & 56.47$\pm$8.45 & 81.82$\pm$0.00 & 0.606 & 0.697 & 0.727 & 0.606 & 0.515 & 0.455 \\
& Claude-Code+Claude-Sonnet-4.5 & 53.23$\pm$7.60 & \textbf{90.91$\pm$9.09} & \textbf{0.667} & 0.697 & 0.727 & \textbf{0.667} & \textbf{0.636} & \textbf{0.636} \\
& Codex+GPT-5.2 & 43.83$\pm$7.15 & 78.79$\pm$13.89 & 0.576 & 0.727 & \textbf{0.818} & 0.576 & 0.424 & 0.364 \\
\midrule
\multirow{4}{*}{\shortstack{Object\\Identification}}
& ParaView-MCP+GPT-5.2 & 32.43$\pm$8.13 & 60.49$\pm$14.02 & \textbf{0.420} & \textbf{0.654} & \textbf{0.778} & \textbf{0.420} & 0.185 & 0.074 \\
& ParaView-MCP+Claude-Sonnet-4.5 & 42.77$\pm$1.53 & \textbf{92.59$\pm$3.70} & 0.259 & 0.358 & 0.407 & 0.259 & 0.160 & 0.111 \\
& Claude-Code+Claude-Sonnet-4.5 & 44.43$\pm$2.30 & 88.89$\pm$6.42 & 0.358 & 0.556 & 0.741 & 0.358 & 0.160 & \textbf{0.148} \\
& Codex+GPT-5.2 & \textbf{44.50$\pm$4.66} & 92.59$\pm$7.41 & \textbf{0.420} & 0.580 & 0.630 & \textbf{0.420} & \textbf{0.259} & \textbf{0.148} \\
\bottomrule
\end{tabular}
\end{adjustbox}
\end{table*}

\begin{table*}[!htb]
\centering
\small
\caption{Benchmark performance of Claude Code on the bioimage visualization task suite of SciVisAgentBench using {\bf GPT-5.2} as the LLM judge for evaluation. 
Scores and completion rates are reported as mean$\pm$std across three repeated trials. 
}
\label{tab:main_results_bioimage_gpt52}
\vspace{-0.1in}
\begin{adjustbox}{width=\textwidth}
\begin{tabular}{lcccccccc}
\toprule
Setting & Overall Score $\uparrow$ & Completion Rate $\uparrow$ & pass@1 $\uparrow$ & pass@2 $\uparrow$ & pass@3 $\uparrow$ & pass$\textasciicircum1$ $\uparrow$ & pass$\textasciicircum2$ $\uparrow$ & pass$\textasciicircum3$ $\uparrow$ \\
\midrule
Claude-Code+Sonnet-4.5 (without skill) 
& 53.23$\pm$7.60 & 90.91$\pm$9.09 & 0.667 & 0.697 & 0.727 & 0.667 & 0.636 & 0.636 \\

Claude-Code+Sonnet-4.5 (with skill) 
& 59.50$\pm$3.54 & \textbf{96.97$\pm$5.25} & 0.727 & 0.848 & 0.909 & 0.727 & 0.606 & 0.545 \\

Claude-Code+Opus-4.6 (without skill) 
& 57.57$\pm$5.16 & 93.94$\pm$5.25 & 0.697 & 0.758 & 0.818 & 0.697 & 0.636 & 0.636 \\

Claude-Code+Opus-4.6 (with skill) 
& \textbf{70.90$\pm$3.31} & 93.94$\pm$5.25 & \textbf{0.848} & \textbf{0.879} & \textbf{0.909} & \textbf{0.848} & \textbf{0.818} & \textbf{0.818} \\

\bottomrule
\end{tabular}
\end{adjustbox}
\end{table*}







\begin{table*}[!htb]
\centering
\scriptsize
\caption{\hot{Representative qualitative outcomes for ParaView visualization tasks. Each row compares the expert-defined ground truth with outputs from ParaView-MCP, ChatVis, Claude Code, and Codex. Scores below the agent outputs are vision-based scores from LLM judges, computed by summing the goal-level scores reported in Table~\ref{tab:qualitative-combined}. These scores are only part of the final benchmark scores.}}
\label{tab:qualitative-outcome-examples}
\vspace{-0.08in}
\setlength{\tabcolsep}{2pt}

\begin{adjustbox}{width=0.98\textwidth}
\begin{tabular}{|c|c|c|c|c|c|}
\hline
\textbf{Case Name} & \textbf{Ground Truth} & \textbf{ParaView-MCP} & \textbf{ChatVis} & \textbf{Claude Code} & \textbf{Codex} \\
\hline

\texttt{mhd-turbulence\_streamline}
&
\begin{tabular}{@{}c@{}}
\includegraphics[width=0.145\textwidth,height=0.13\textwidth,keepaspectratio]{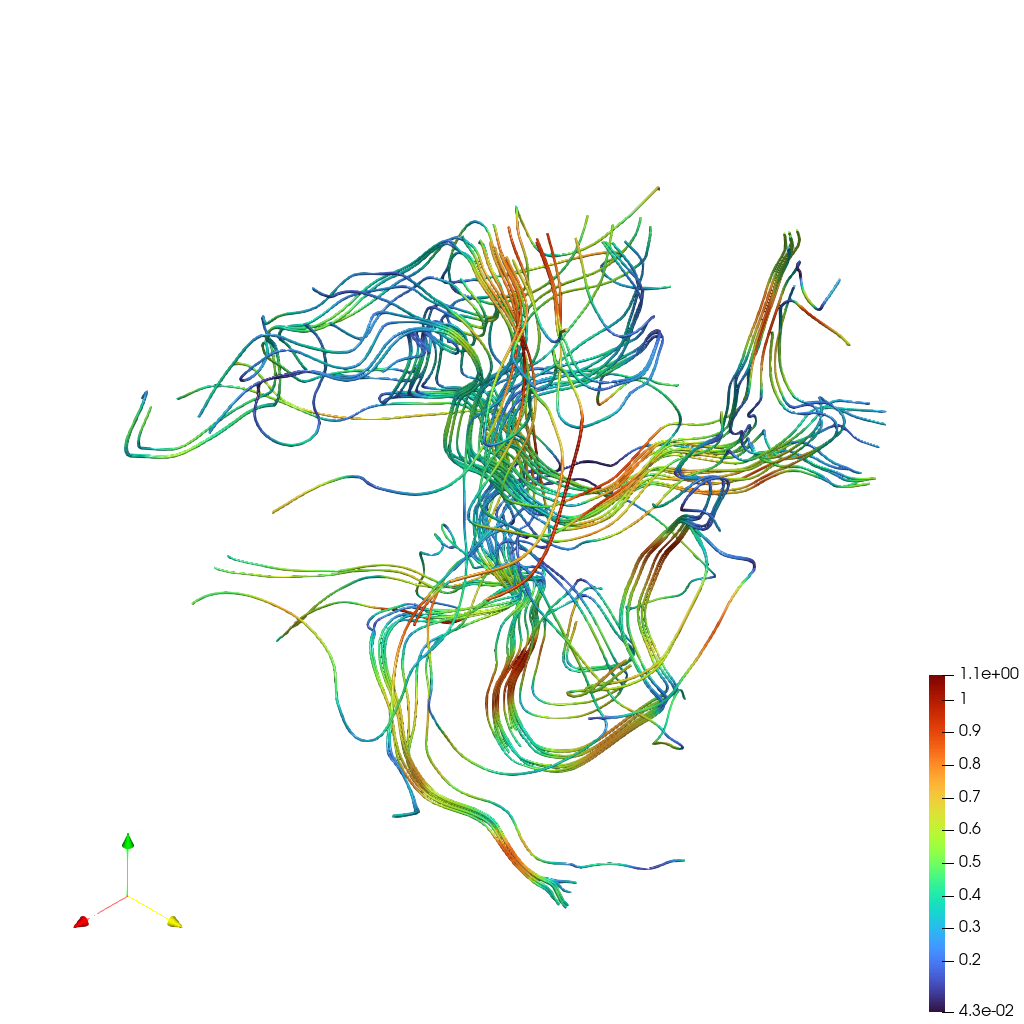}
\end{tabular}
&
\begin{tabular}{@{}c@{}}
\includegraphics[width=0.145\textwidth,height=0.13\textwidth,keepaspectratio]{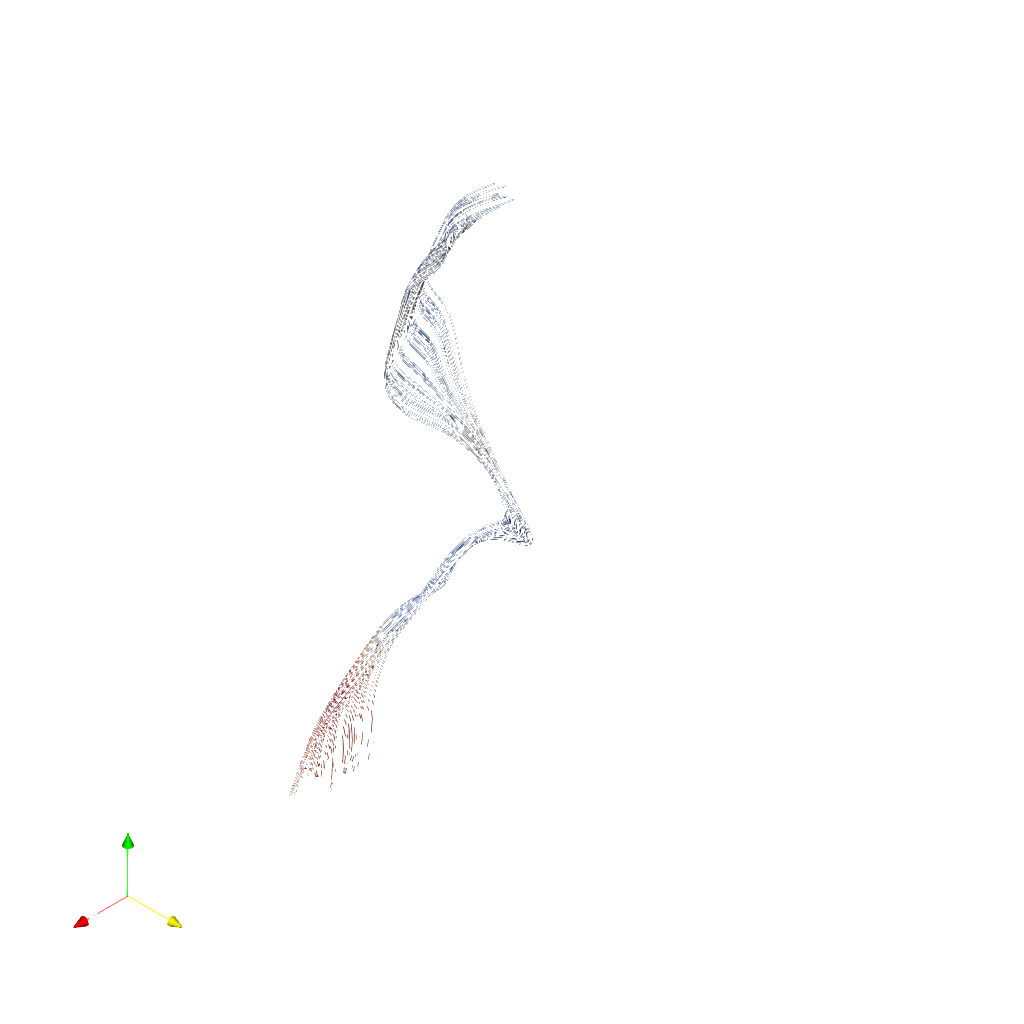}\\[-0.02in]
\scriptsize Score: 8/40 (20\%)
\end{tabular}
&
\begin{tabular}{@{}c@{}}
\includegraphics[width=0.145\textwidth,height=0.13\textwidth,keepaspectratio]{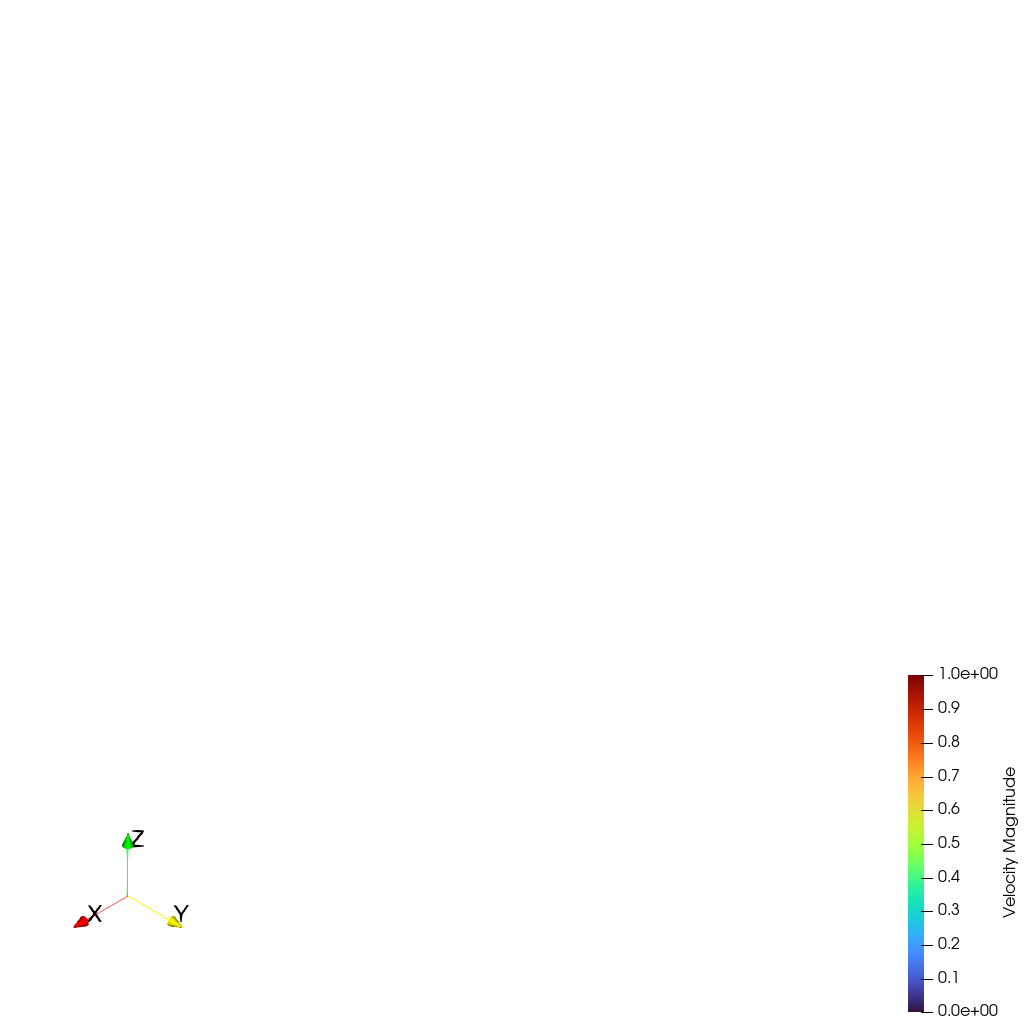}\\[-0.02in]
\scriptsize Score: 0/40 (0\%)
\end{tabular}
&
\begin{tabular}{@{}c@{}}
\includegraphics[width=0.145\textwidth,height=0.13\textwidth,keepaspectratio]{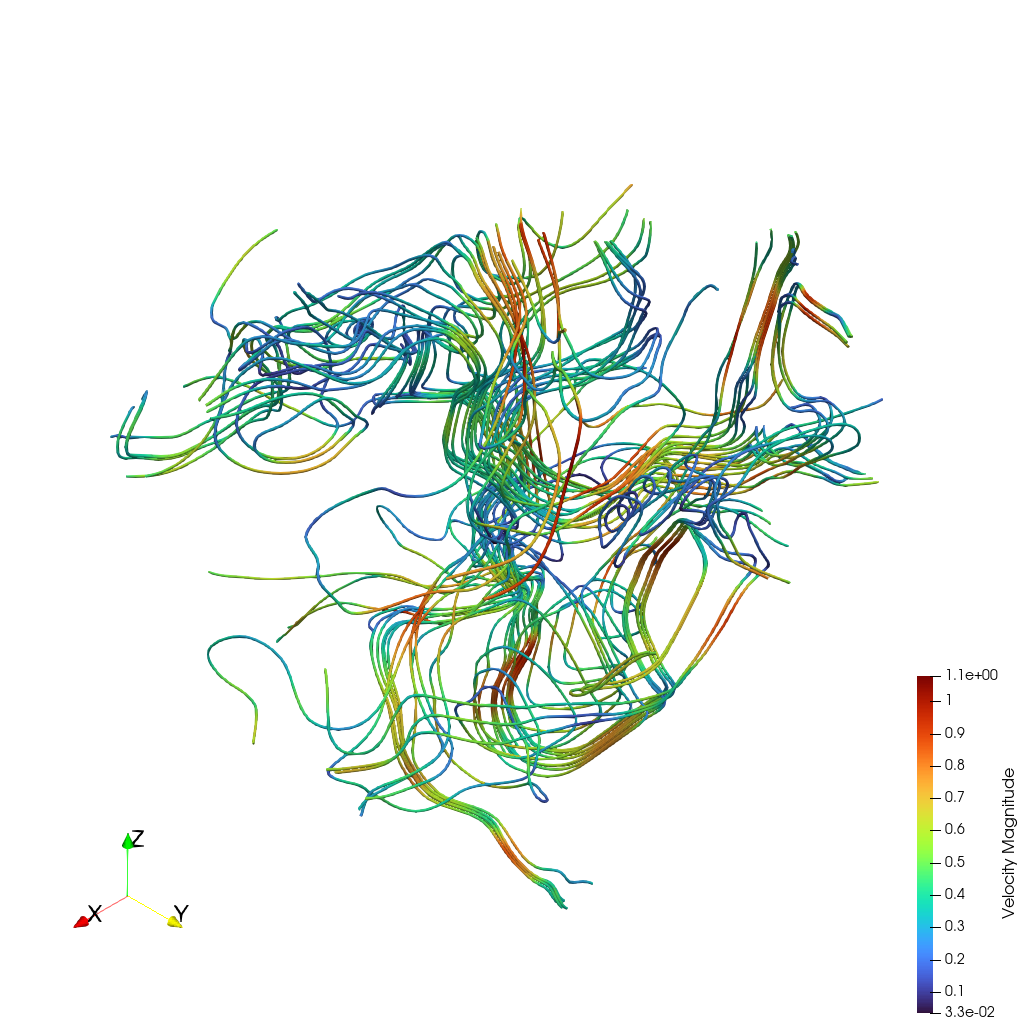}\\[-0.02in]
\scriptsize Score: 35/40 (87.5\%)
\end{tabular}
&
\begin{tabular}{@{}c@{}}
\includegraphics[width=0.145\textwidth,height=0.13\textwidth,keepaspectratio]{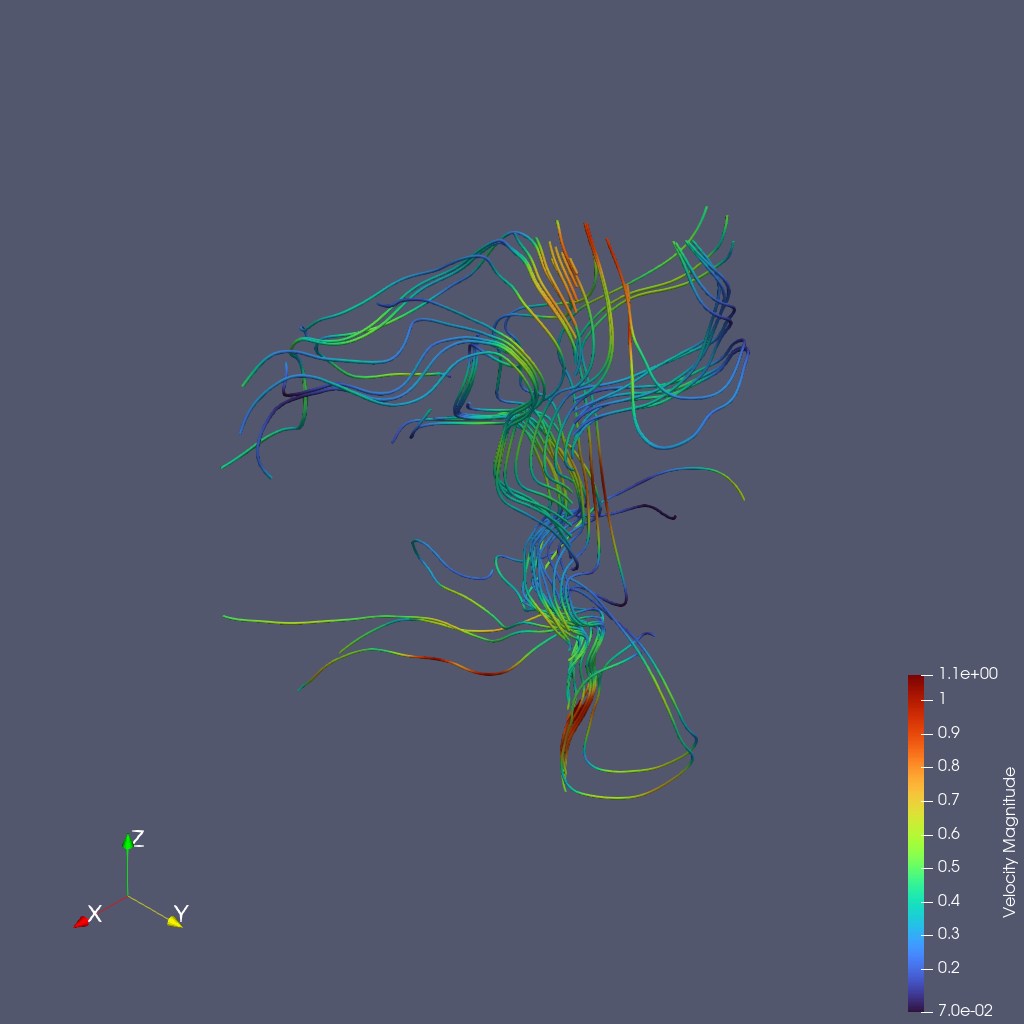}\\[-0.02in]
\scriptsize Score: 24/40 (60\%)
\end{tabular}
\\
\hline

\texttt{render-histogram}
&
\begin{tabular}{@{}c@{}}
\includegraphics[width=0.145\textwidth,height=0.13\textwidth,keepaspectratio]{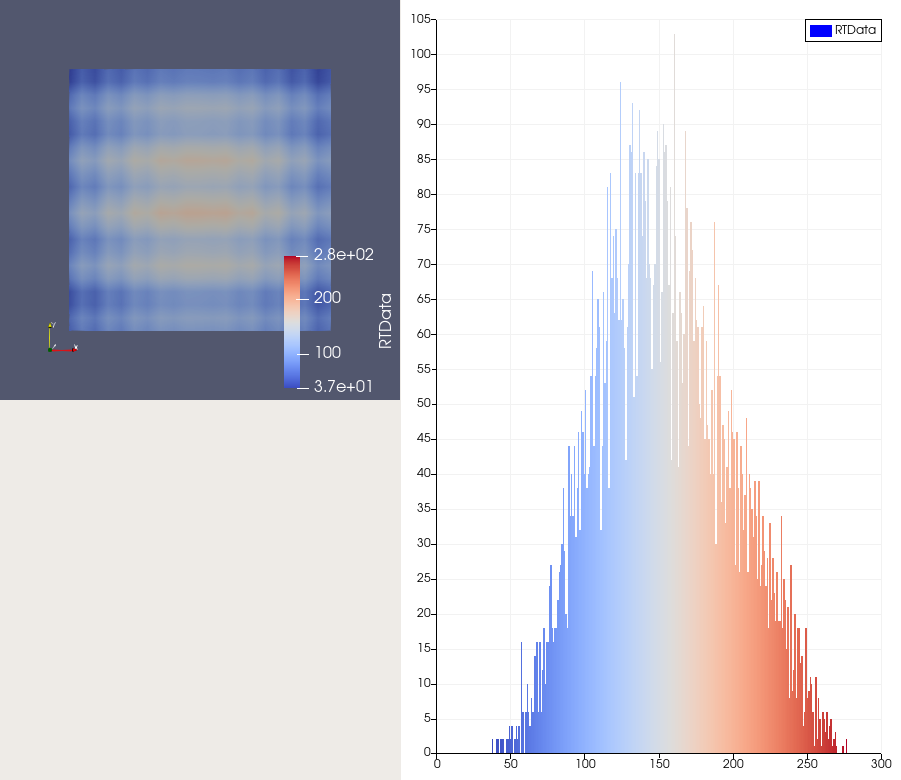}
\end{tabular}
&
\begin{tabular}{@{}c@{}}
\includegraphics[width=0.145\textwidth,height=0.13\textwidth,keepaspectratio]{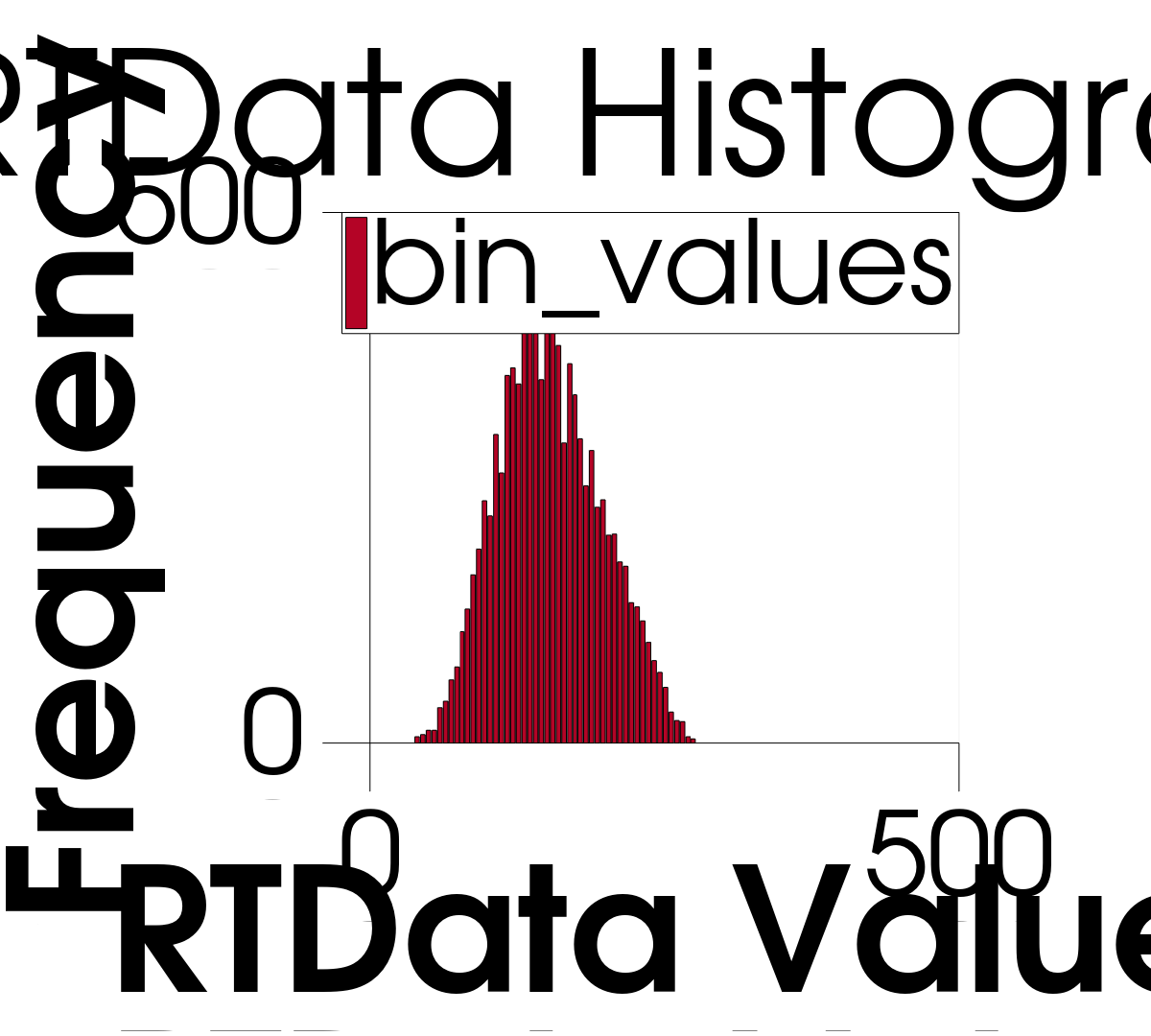}\\[-0.02in]
\scriptsize Score: 17/40 (42.5\%)
\end{tabular}
&
\begin{tabular}{@{}c@{}}
\includegraphics[width=0.145\textwidth,height=0.13\textwidth,keepaspectratio]{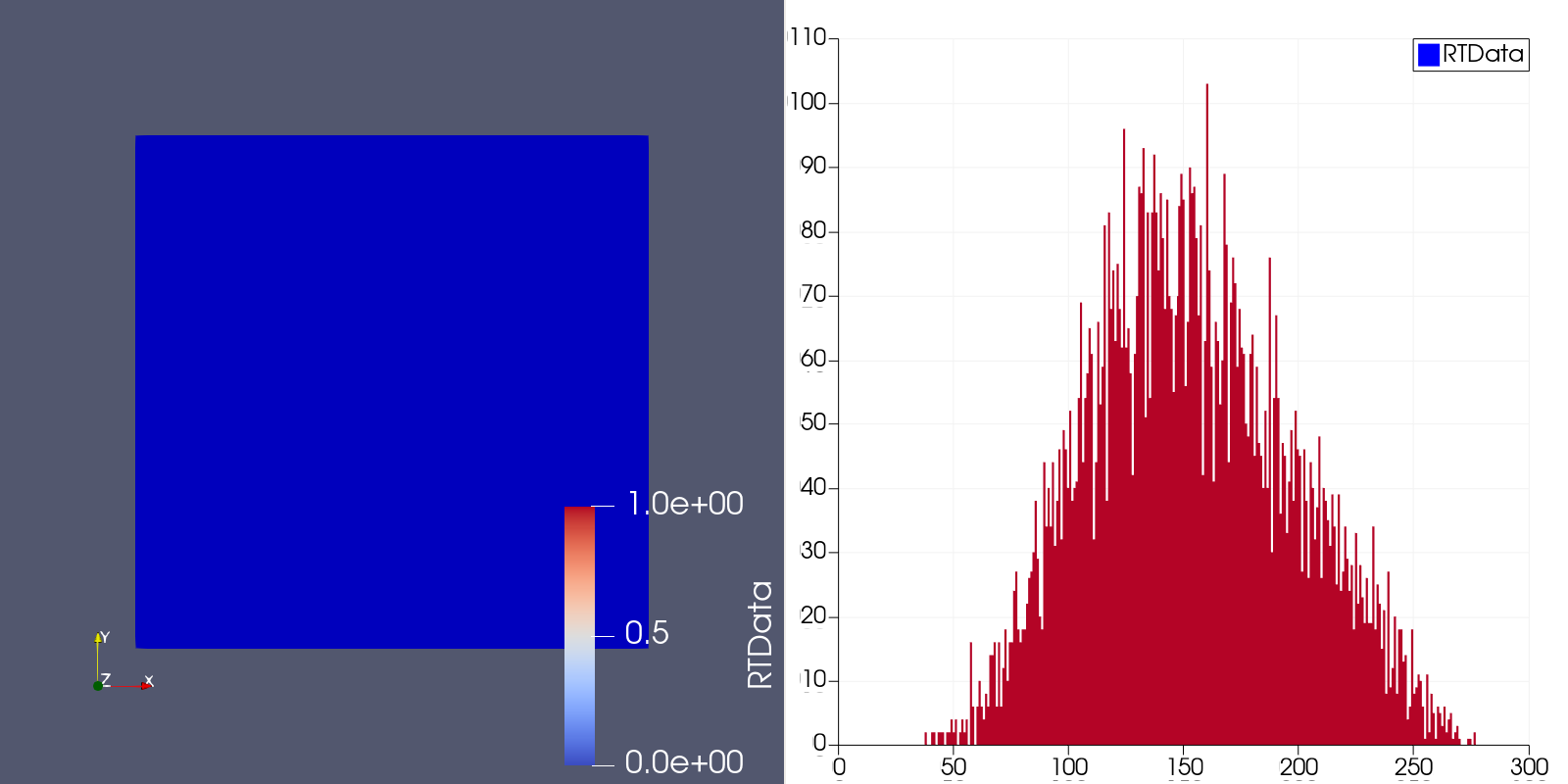}\\[-0.02in]
\scriptsize Score: 25/40 (62.5\%)
\end{tabular}
&
\begin{tabular}{@{}c@{}}
\includegraphics[width=0.145\textwidth,height=0.13\textwidth,keepaspectratio]{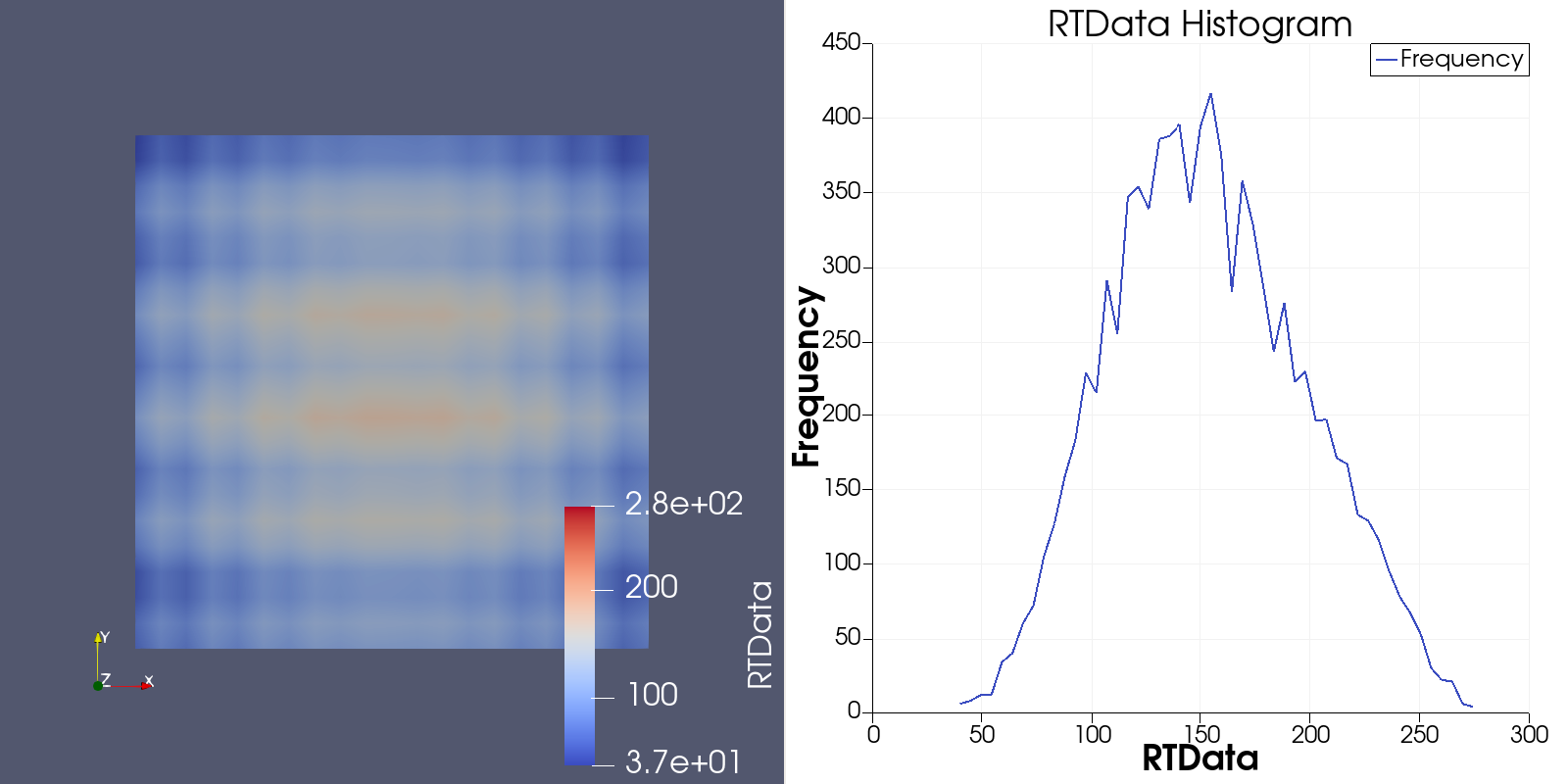}\\[-0.02in]
\scriptsize Score: 29/40 (72.5\%)
\end{tabular}
&
\begin{tabular}{@{}c@{}}
\includegraphics[width=0.145\textwidth,height=0.13\textwidth,keepaspectratio]{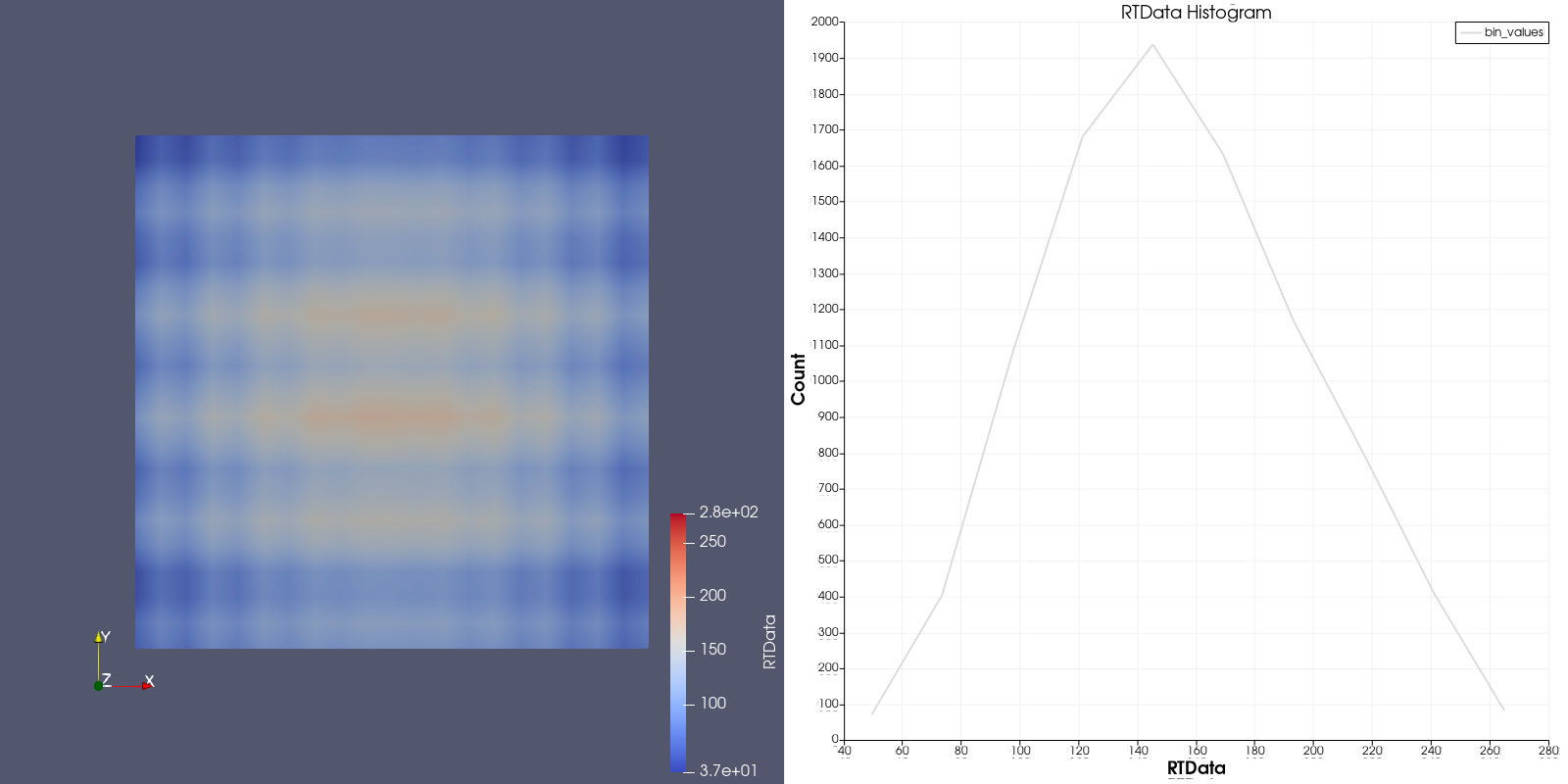}\\[-0.02in]
\scriptsize Score: 28/40 (70\%)
\end{tabular}
\\
\hline

\texttt{lobster}
&
\begin{tabular}{@{}c@{}}
\includegraphics[width=0.145\textwidth,height=0.13\textwidth,keepaspectratio]{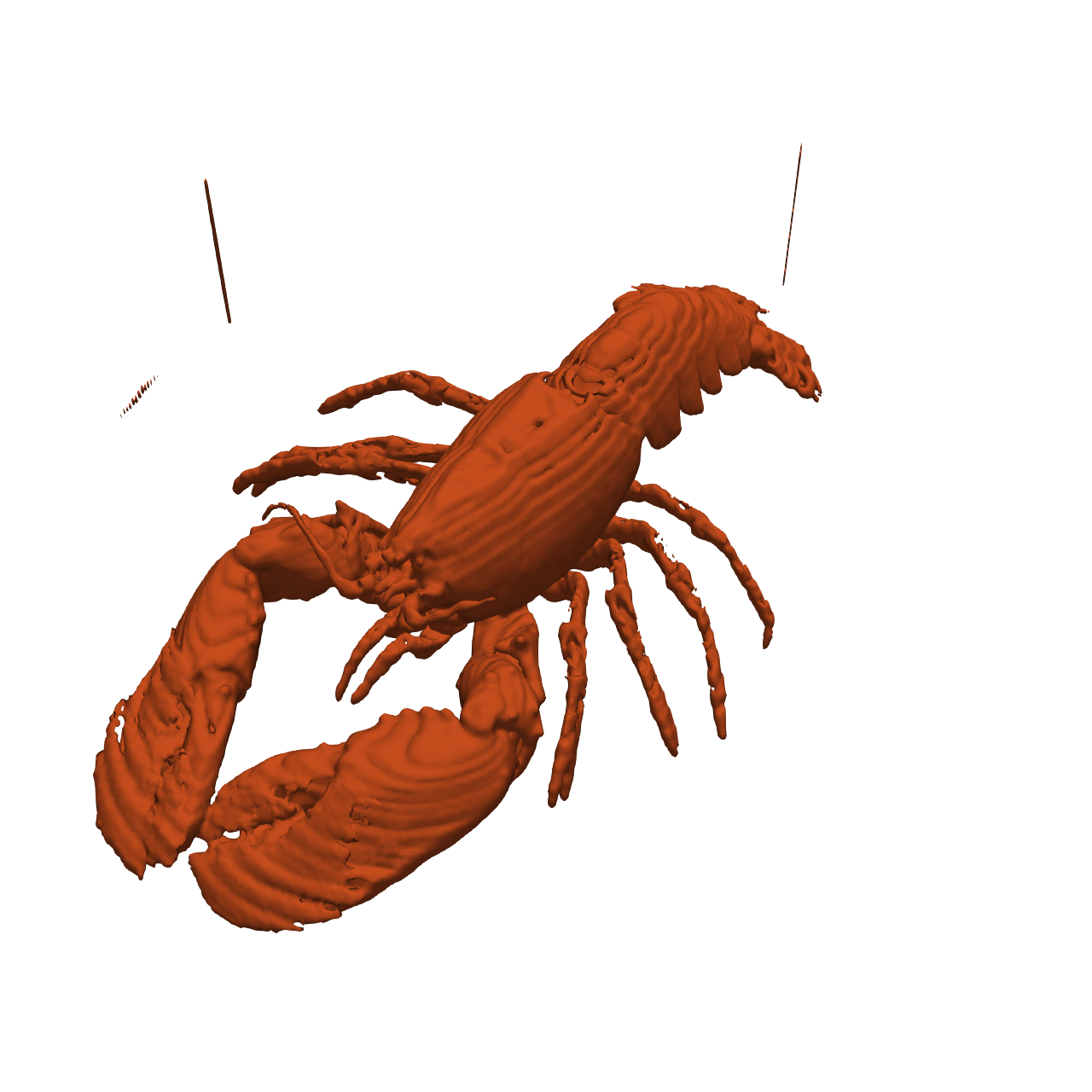}
\end{tabular}
&
\begin{tabular}{@{}c@{}}
\includegraphics[width=0.145\textwidth,height=0.13\textwidth,keepaspectratio]{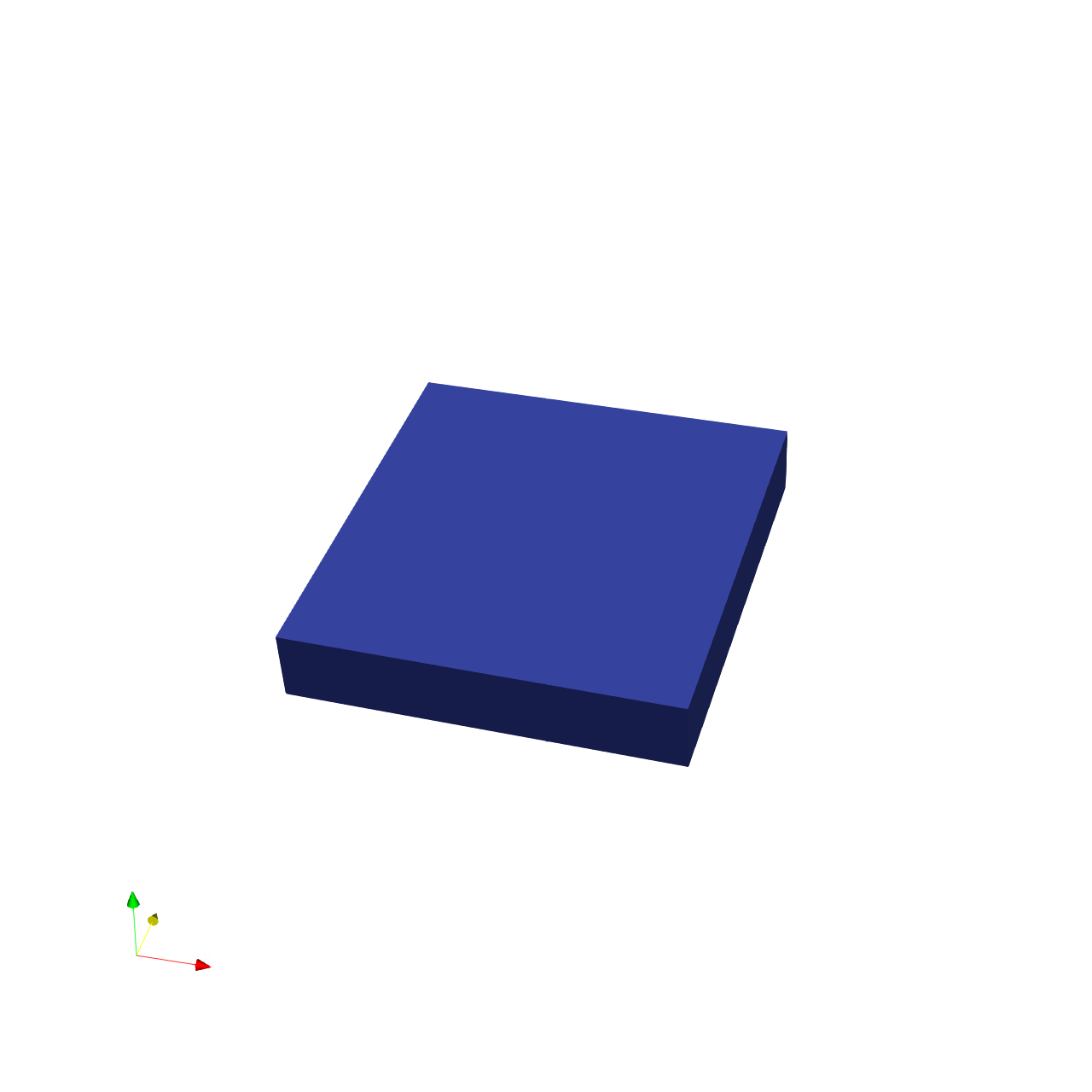}\\[-0.02in]
\scriptsize Score: 0/30 (0\%)
\end{tabular}
&
\begin{tabular}{@{}c@{}}
\includegraphics[width=0.145\textwidth,height=0.13\textwidth,keepaspectratio]{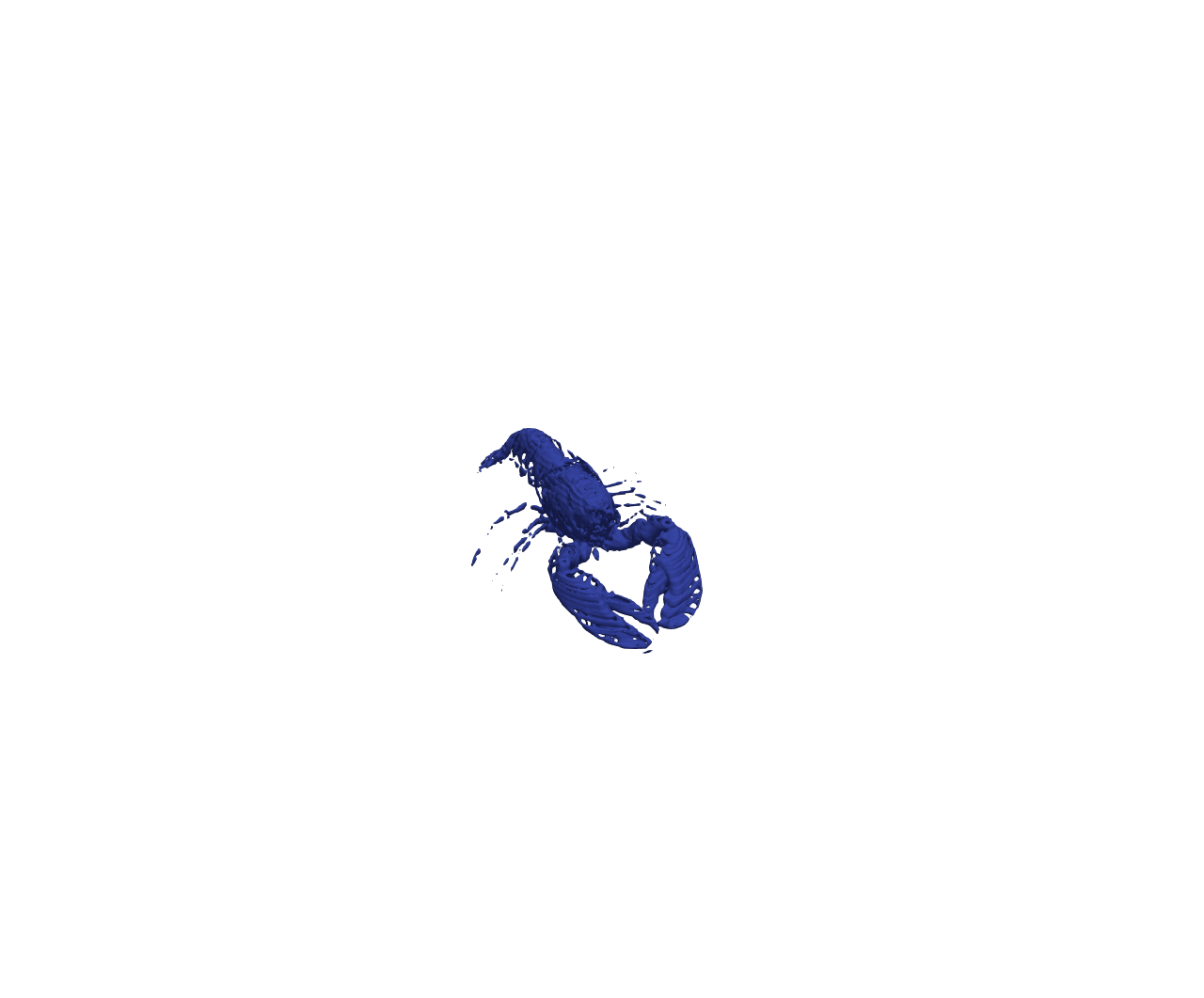}\\[-0.02in]
\scriptsize Score: 6/30 (20\%)
\end{tabular}
&
\begin{tabular}{@{}c@{}}
\includegraphics[width=0.145\textwidth,height=0.13\textwidth,keepaspectratio]{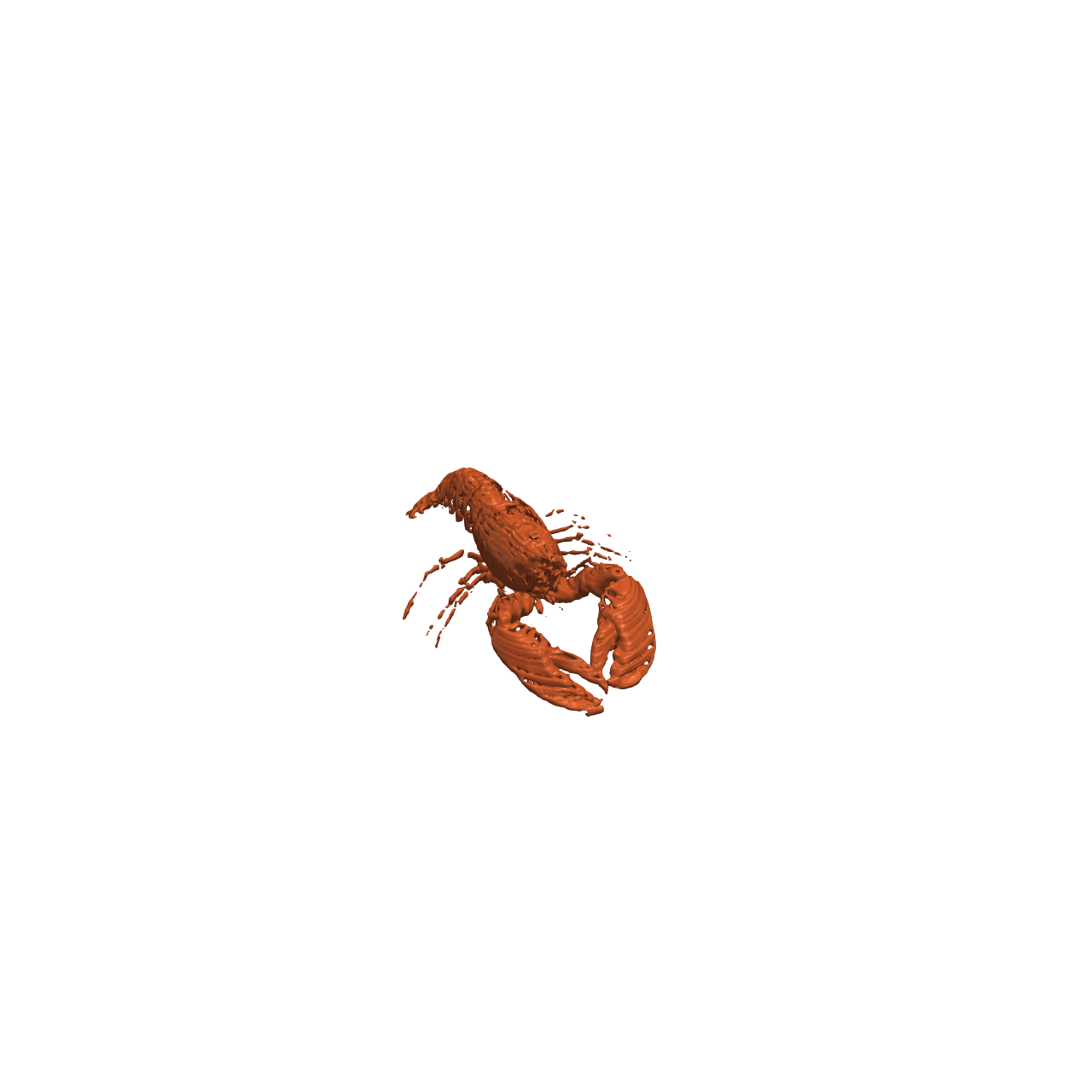}\\[-0.02in]
\scriptsize Score: 19/30 (63.3\%)
\end{tabular}
&
\begin{tabular}{@{}c@{}}
\includegraphics[width=0.145\textwidth,height=0.13\textwidth,keepaspectratio]{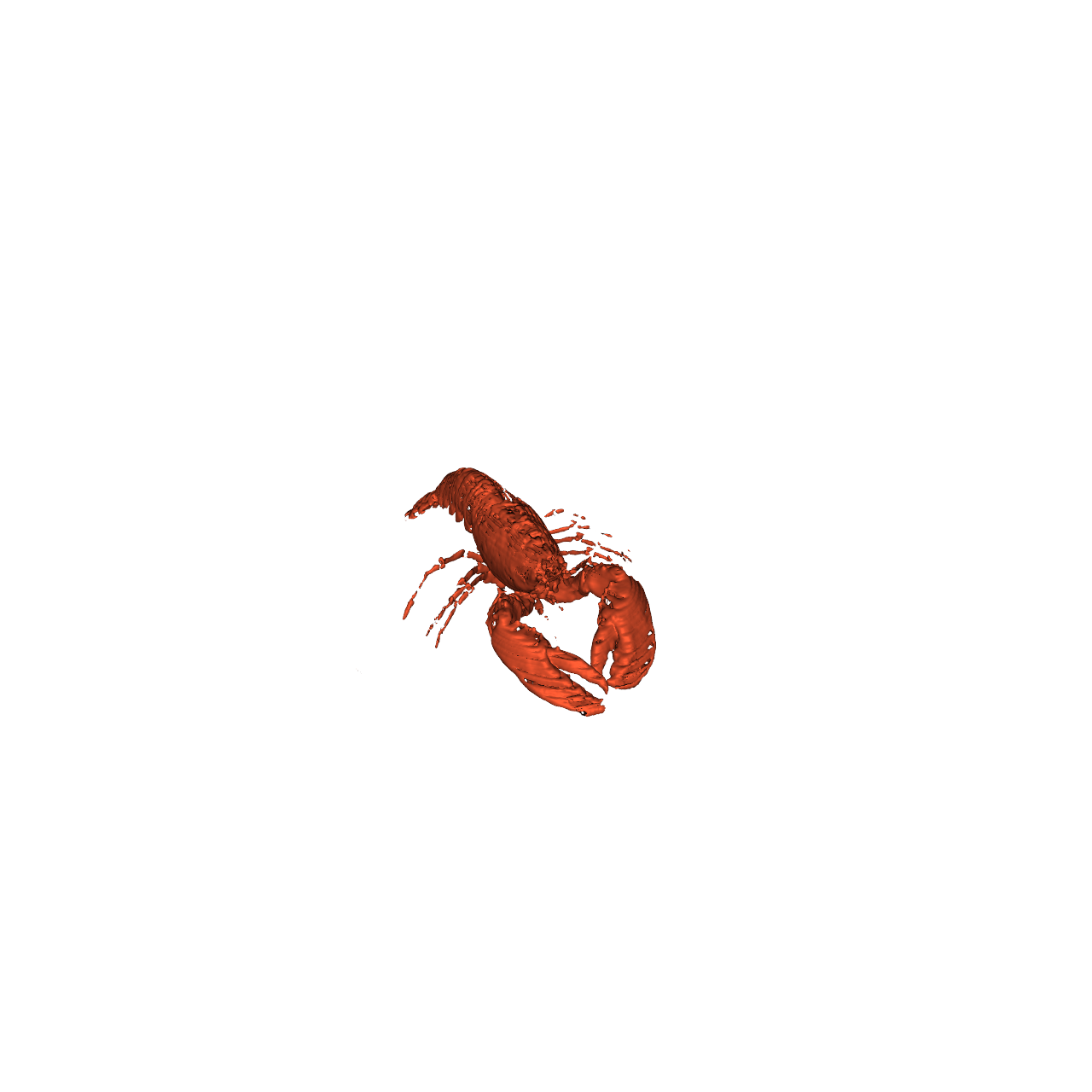}\\[-0.02in]
\scriptsize Score: 17/30 (56.7\%)
\end{tabular}
\\
\hline

\end{tabular}
\end{adjustbox}
\end{table*}

\begin{table*}[!htb]
\centering
\scriptsize
\caption{\hot{Task descriptions, goal-level evaluation criteria, and per-goal vision-based rubric scores for the examples in Table~\ref{tab:qualitative-outcome-examples}. Each case lists the benchmark instruction, the scientific or visualization property assessed by each goal, and the corresponding scores for ParaView-MCP, ChatVis, Claude Code, and Codex. The total row sums all vision-based goal scores and explains the score shown below each image in Table~\ref{tab:qualitative-outcome-examples}. This total is only one component of the final benchmark score.}}
\label{tab:qualitative-combined}
\vspace{-0.08in}
\setlength{\tabcolsep}{2pt}
\renewcommand{\arraystretch}{1.12}

\begin{adjustbox}{width=\textwidth}
\begin{tabular}{
p{0.46\textwidth}
c
p{0.37\textwidth}
c
c
c
c
}
\toprule
\textbf{Case Name \& Task Description} &
\textbf{Goal} &
\textbf{Criterion} &
\textbf{ParaView-MCP} &
\textbf{ChatVis} &
\textbf{Claude Code} &
\textbf{Codex} \\
\midrule

\multirow{5}{=}{
\begin{minipage}[c]{\linewidth}
\makebox[\linewidth][c]{\texttt{\scriptsize mhd-turbulence\_streamline}}

\vspace{0.04in}
\raggedright
Load the $128^3$ VTI MHD turbulence velocity field; generate 3D streamlines from a $z$-axis line source at $x{=}64$, $y{=}64$ with 50 seed points; color streamlines by velocity magnitude using the Turbo colormap; apply a Tube filter with radius 0.3; add a ``Velocity Magnitude'' color bar; use a white background and the specified isometric camera; and save the rendered visualization and ParaView state or script files.
\end{minipage}
}
& G1 &
\textbf{Overall Visualization Goal:} Does the result match the ground-truth streamline visualization of the MHD turbulence velocity field?
& 2/10 & 0/10 & 9/10 & 6/10 \\
& G2 &
\textbf{Streamline Patterns:} Do the streamlines show similar flow patterns and structures as the ground truth?
& 2/10 & 0/10 & 9/10 & 6/10 \\
& G3 &
\textbf{Streamline Coverage:} Is the spatial distribution and density of streamlines similar to the ground truth?
& 1/10 & 0/10 & 9/10 & 5/10 \\
& G4 &
\textbf{Color Mapping:} Is the color distribution along streamlines visually similar to the ground truth?
& 3/10 & 0/10 & 8/10 & 7/10 \\
& \textbf{Total} &
& \textbf{8/40} & \textbf{0/40} & \textbf{35/40} & \textbf{24/40} \\

\midrule

\multirow{5}{=}{
\begin{minipage}[c]{\linewidth}
\makebox[\linewidth][c]{\texttt{\scriptsize render-histogram}}

\vspace{0.04in}
\raggedright
Create a wavelet object; render it as an RTData-colored surface with a visible color bar; rescale colors to the data range using the Cool-to-Warm colormap; split the view horizontally; create an RTData histogram on the right using the same colormap; and save one screenshot plus ParaView state or script files.
\end{minipage}
}
& G1 &
\textbf{Wavelet Visualization:} Is the wavelet object properly rendered with RTData coloring and a visible color bar?
& 6/10 & 4/10 & 9/10 & 9/10 \\
& G2 &
\textbf{Split View Layout:} Is the view correctly split with the wavelet visualization on the left and histogram on the right?
& 4/10 & 10/10 & 10/10 & 10/10 \\
& G3 &
\textbf{Histogram Generation:} Is the histogram properly generated from RTData, showing the data distribution?
& 5/10 & 8/10 & 6/10 & 4/10 \\
& G4 &
\textbf{Color Map Consistency:} Are both the wavelet visualization and histogram using the same Cool-to-Warm color map?
& 2/10 & 3/10 & 4/10 & 5/10 \\
& \textbf{Total} &
& \textbf{17/40} & \textbf{25/40} & \textbf{29/40} & \textbf{28/40} \\

\midrule

\multirow{4}{=}{
\begin{minipage}[c]{\linewidth}
\makebox[\linewidth][c]{\texttt{\scriptsize lobster}}
\vspace{0.04in}
\raggedright
Load the raw Lobster CT dataset with the correct uint8 type, little-endian byte order, $301\times324\times56$ extent, and $1\times1\times1.4$ spacing; extract a specimen-boundary isosurface; color it red-orange; answer the walking-leg multiple-choice question; use a white background, optimal view, no color bar or axes, and the specified camera; and save the visualization, text answer, and ParaView state or script files.
\end{minipage}
}
& G1 &
\textbf{Overall Goal:} Does the visualization clearly show the structure and details of the lobster?
& 0/10 & 3/10 & 5/10 & 4/10 \\[0.2in]
& G2 &
\textbf{Boundary Clarity:} Are the surface details and boundaries of the lobster well-defined?
& 0/10 & 3/10 & 6/10 & 6/10 \\[0.2in]
& G3 &
\textbf{Correct Color:} Does the color of the lobster mimic a real lobster, i.e., red-orange?
& 0/10 & 0/10 & 8/10 & 7/10 \\[0.2in]
& \textbf{Total} &
& \textbf{0/30} & \textbf{6/30} & \textbf{19/30} & \textbf{17/30} \\

\bottomrule
\end{tabular}
\end{adjustbox}
\end{table*}

\vspace{-0.05in}
\hot{\section{Vision-Based Outcome Evaluation Examples with LLM Judges}
\label{appendix:qualitative-examples}

We provide three representative qualitative examples to make the benchmark scores more interpretable and to address concerns that aggregate numerical metrics alone may obscure scientific failure modes. For each case, we compare the expert-defined ground-truth visualization with outputs from different agents and report the vision-based score assigned by the LLM judge, including its goal-level rubric breakdown. These scores are not the final benchmark scores reported in the main result tables. The final benchmark score also incorporates deterministic components, such as output-compliance checks, token usage, execution time, multiple-choice answers, and other case-specific validators, as applicable. These examples show that differences in the vision-based component are not merely low-level image-similarity artifacts: they often correspond to concrete visualization or scientific errors, such as missing flow structures, incorrect chart construction, weak boundary extraction, or mismatched visual encodings.

Claude Sonnet 4.5 is the primary agent model for ParaView-MCP, ChatVis, and Claude Code, whereas Codex uses GPT-5.2. GPT-5.2 evaluates all outcomes shown in this appendix. In Tables~\ref{tab:qualitative-outcome-examples} and~\ref{tab:qualitative-combined}, each vision-based score is the sum of the per-goal rubric scores for that case; each goal is scored on a 0-10 scale, and the denominator is the maximum possible score for all applicable goals. The full set of all 108 cases and outcomes from all tested agents is available on the project page.}

\hot{
\vspace{-0.05in}
\section{Further Analysis on Coding Agents with Skills}
\label{appendix:agent_skill}

Coding agents often fail due to frictions in tool setup and use. In the main benchmark test, we provided no additional tool instructions to obtain an unbiased baseline. However, such frictions can be mitigated by using agent skills~\cite{anthropic-as}: lightweight instruction documents with working code snippets and tool-specific guidance that help agents focus on the task rather than environment exploration.

We conduct a focused follow-up study on bioimage visualization tasks, in which coding agents underperformed purpose-built MCP-based agents. To preserve benchmark integrity, the skill was developed without exposing the agent to benchmark tasks. Instead, we asked the agent to inspect the napari MCP agent, distill its capabilities into a reusable napari skill, and test the skill in the provided environment. The resulting skill captured practical implementation details, version-specific constraints, and common failure-avoidance patterns.

For a direct comparison, we evaluated Claude Code with and without the skill using Sonnet-4.5 and Opus-4.6. GPT-5.2 was used as the judge to avoid using Opus-4.6 both as the task agent and evaluator. As shown in Table~\ref{tab:claude_code_bioimage_summary}, skills improve performance for both models and reduce token consumption by roughly half. Opus-4.6 consistently achieves stronger results with fewer tokens, suggesting that stronger models can solve the task more efficiently with fewer interaction steps. Full bioimage results are provided in Table~\ref{tab:main_results_bioimage_gpt52}. 

The results show that Claude Code with napari skill achieved a score of 70.9 compared to 57.6 without skill, a 23\% improvement, while simultaneously reducing token consumption by approximately 50\%. This suggests that the most effective architecture for SciVis agents may center on flexible reasoning engines paired with modular, domain-specific knowledge that provides tool-use patterns without constraining exploratory capabilities. 
A broader follow-up study, SciVisAgentSkills~\cite{ai2026scivisagentskills}, further extends this analysis to reusable skills across multiple SciVis tools and task suites.}

\begin{table*}[htb]
\centering
\small
\caption{Claude Code results on the bioimage visualization task suite of SciVisAgentBench with different backbone models and skill settings. 
{\bf GPT-5.2} is used as the LLM judge for evaluation, as Opus-4.6 is used as the task agent. 
Values are mean$\pm$std across three repeated trials. 
}
\label{tab:claude_code_bioimage_summary}
\vspace{-0.1in}
\begin{tabular}{lcccc}
\toprule
Setting & Overall Score $\uparrow$ & Completion Rate $\uparrow$ & Input Tokens $\downarrow$ & Output Tokens $\downarrow$ \\
\midrule
Claude-Code+Sonnet-4.5 (without skill)
& 53.23$\pm$7.60 & 90.91$\pm$9.09 & 8.60M$\pm$0.17M & 125.66K$\pm$2.43K \\
Claude-Code+Sonnet-4.5 (with skill)
& 59.50$\pm$3.54 & \textbf{96.97$\pm$5.25} & 4.31M$\pm$3.83M & 53.14K$\pm$41.02K \\
Claude-Code+Opus-4.6 (without skill)
& 57.57$\pm$5.16 & 93.94$\pm$5.25 & 6.67M$\pm$1.17M & 97.98K$\pm$16.93K \\
Claude-Code+Opus-4.6 (with skill) 
& \textbf{70.90$\pm$3.31} & 93.94$\pm$5.25 & \textbf{2.67M$\pm$0.05M} & \textbf{44.81K$\pm$0.69K} \\
\bottomrule
\end{tabular}
\end{table*}

\end{document}